# Enhancing Tree Species Classification: Insights from YOLOv8 and Explainable AI Applied to TLS Point Cloud Projections

**Adrian Straker[1*], Paul Magdon[1], Marco Zullich[2,3], Maximilian Freudenberg[4], Christoph Kleinn[5], Johannes Breidenbach[6], Stefano Puliti[6], Nils Nölke[5]**

*[1] Faculty of Resource Management, University of Applied Sciences and Art (HAWK), Göttingen, Büsgenweg 1a, 37077 Göttingen, Germany*

*[2] Department of Artificial Intelligence, University of Groningen, Nijenborgh 9, 9747 AG Groningen, the Netherlands*

*[3] Faculty of Technology and Policy Management, Delft University of Technology, Jaffalaan 5, 2628 BX Delft, the Netherlands*

*[4] Neural Data Science Group, University of Göttingen, Goldschmidtstraße 3, 37077 Göttingen, Germany*

*[5] Forest Inventory and Remote Sensing, University of Göttingen, Faculty of Forest Sciences and Forest Ecology, Burckhardt-Institute, Büsgenweg 5, 37077 Göttingen, Germany*

*[6] Division of Forest and Forest Resources, National Forest Inventory, Norwegian Institute of Bioeconomy Research (NIBIO), Høgskolveien 8, 1433 Ås, Norway*

*Corresponding author: Adrian Straker; Tel: +49 551/5032-73127; Email: adrian.straker@stud.uni-goettingen.de

## Abstract

Classifying tree species has been a core research area in forest remote sensing for decades. The application of deep learning models using data from novel terrestrial laser scanning (TLS) sensors achieves high accuracies but the classification decision processes of these models remain unclear. Aiming to advance research in the field of interpretability of deep learning models for tree species classification using TLS 3D point clouds we present insights in the classification capabilities of YOLOv8 through a new framework which enables systematic analysis of saliency maps derived from CAM (Class Activation Mapping). In this context, we adopted Finer-CAM, which highlights image regions in 2D side-view projections (images) of TLS 3D point clouds that contribute to the classification of a target species, but are uncommon in similar looking contrastive tree species. To investigate the contribution of structural tree features (stem, crown, crown sub-segments) to the classification decisions of YOLOv8 models, we then link regions with high saliency derived from the application of Finer-CAM to segments of 2D side-view images that correspond to structural tree features. Using TLS 3D point clouds from 2,445 trees across seven European tree species, we trained and validated five YOLOv8 models with cross-validation, reaching a mean accuracy of 96% (SD = 0.24%) when applied to the test data. Our results demonstrate that Finer-CAM can be considered faithful in identifying discriminative regions that distinguish target tree species from their most similar counterparts. This renders Finer-CAM suitable for enhancing the interpretability of the tree species classification models.

Analysis of 630 Finer-CAM saliency maps indicate that the models primarily rely on image regions associated with tree crowns  for species classification. While this result is pronounced in *Silver Birch*, *European Beech*, *English oak*, and *Norway Spruce*, image regions associated with stems contribute more frequently to the differentiation of *European ash*, *Scots pine*, and *Douglas-fir*. We further demonstrate that the visibility of detailed structural tree features in the 2D side-view images enhances the discriminative performances of the models, indicating YOLOv8's capabilities to leverage detailed TLS point cloud representations. Our results represent a first step toward enhancing the understanding of the classification decision processes of tree species classification models, aiding in the identification of data set and model limitations, biases, and building confidence in model predictions.

## Introduction

Tree species classification from remote sensing data has long been a challenging task. In addition to image-based data sets, 3D point clouds have been increasingly used in recent years. Terrestrial laser scanning (TLS) has emerged as a useful remote sensing tool for capturing detailed 3D information of individual trees. In recent years, deep learning models, particularly convolutional neural networks (CNNs), have become state-of-the-art for automatically classifying tree species using TLS data and data from other proximal sensors (Puliti et al. 2025). Theclassification approaches often leverage 2D projections of TLS point clouds (Mizoguchi et al. 2017, Seidel et al. 2021, Allen et al. 2023, Puliti et al. 2025) or operate directly on the 3D point cloud data itself (Liu et al. 2021, Chen et al. 2021, Liu et al. 2022, Puliti et al. 2025). Using the FOR-species20k data set, Puliti et al. (2025) benchmarked several tree species classification approaches and found multi-view CNNs operating on 2D projections (images) of proximal laser scanning 3D point clouds slightly outperform native 3D CNNs. Example multi-view CNNs are DetailView, YOLOv5, and SimpleView, which achieved accuracies of 79.5 %, 77.9 % and 76.2 %, respectively. In the context of forest remote sensing, YOLO-style models have been widely used for tree crown segmentation from laser scanning data (Straker et al. 2023) or from aerial imagery (Sun et al. 2025), as well as to detect tree species, health, and crown damages (Puliti & Astrup 2022)

The training of novel CNN-based tree species classification models on publicly available tree 3D point cloud data sets suggests that their operational deployment on real-world data is feasible. However, while these recent advancements show that tree species classification using TLS data seems to work well, little is known about how CNN-based tree species classification models take their decisions and which information in the TLS datasets is being used for separating the tree species. In most studies so far, only standard evaluation metrics (e.g., recall, precision, F1-score and overall accuracy) are reported (e.g., Seidel et al. 2021, Allen et al. 2023, Puliti et al. 2025). These metrics focus on the performance outcome of a model but do not provide information on a model's decision process. Investigating the reasoning of a model for a final prediction is important to improve understanding of model limitations, potential biases, and to gain confidence in a model's predictions (Molnar 2025). Studies from other fields have shown that CNN-based models may learn shortcuts in the provided training data that correspond to data artifacts (e.g., Lapuschkin et al. 2019, Geirhos et al. 2020).

This so-called shortcut learning leads to high model performance in experimental settings but hinders model generalizability when applied to real life data (Hinns & Martens 2025). Consequently, understanding how deep learning models come to their decisions can facilitate the creation of artifact-free labelled training data sets, which in turn hinders short cut learning and in doing so can be used to further improve classification methods (e.g., Lines et al. 2022).

Explainable Artificial Intelligence (XAI) methods provide a valuable framework for understanding how CNN models make predictions. In the context of tree species classification, the application of XAI methods may offer a promising approach to identify the structural features of a tree—such as crown shape, branching patterns, and stem morphology— that most influence model classification decisions. This in turn, can reveal potential model limitations and contribute to building trust in the predictions. One widely used XAI approach is the construction of saliency maps (feature importance), which provide a visualization that highlights image regions most relevant to a model's classification decision. Saliency maps are considered local explanations, since they visually highlight pixels with high discriminative power only for individual inputs. For the creation of saliency maps Zhou et al. (2016) developed Class Activation Mapping (CAM) which generates a weighted sum of the outputs of global average pooling of each feature map at the last convolutional layer to produce saliency maps. Based on this groundwork other approaches like Grad-CAM (Selvaraju et al. 2017), Grad-CAM++ (Chattopadhay et al. 2018), EigenCAM (Muhammad et al. 2020), and Finer-CAM (Zhang et al. 2025) have been developed. In general CAM methods attribute a value of saliency to image regions that contribute to class prediction. However, as models can rely on the same or similar image features for classifying similar classes, these methods are insufficient in identifying features that differentiate similar classes. To generate saliency maps that emphasize discriminative image regions between similar classes Finer-CAM (Zhang et al. 2025) can be applied, as it suppresses features of the class most similar to a target class. To verify the alignment of CAM methods with the model predictive dynamics, faithfulness scores can be assessed (see Bach et al. 2015, Samek et al 2021, Mamalakis et al. 2025). Faithfulness is commonly analysed by the application of least relevant first (LeRF) or most relevant first (MoRF) image perturbation strategies, in which image regions are gradually perturbed based on their saliency assigned by the explanation. The perturbed images are fed into the trained models and changes in model confidence are assessed. Kadir et al. (2023) provide a general overview of different faithfulness estimate metrics and Bücher et al. (2024) tested the effects of various image perturbation strategies on faithfulness metrics.

In the field of forest sciences, CAM methods have been used to understand tree species classification by CNNs using drone-based images (Onishi & Ise 2021, Ma et al. 2024), for the identification of bark features for tree species classification (Kim et al. 2022), for the detection of tree crowns in aerial images (Marvasti-Zadeh et al. 2023), and for the differentiation between leaves and wood in individual tree TLS point clouds (Han & Sánchez-Azofeifa 2022). In the context of tree species classification on TLS point clouds Xi et al. (2020) and Liu et al. (2022) visually interpreted saliency maps to assess tree feature contribution to model classification decisions. However, a comprehensive and systematic analysis of the applicability of saliency maps to elucidate how CNN-based tree species classification models utilizing TLS 3D point clouds arrive at their classification decisions remains lacking. Additionally, the visual interpretation of saliency maps can be influenced by cognitive biases (e.g., Mohseni et al. 2021; Bertrand et al. 2022). Therefore, manually analysing large numbers of saliency maps to identify recurring class specific patterns of highly discriminative image regions is impractical.

To address these limitations, Hinns & Martens (2025) have developed a method that provides semi-global explanations by extracting class specific patterns from saliency map, thereby facilitating the detection of shortcut learning (Geirhos et al. 2020). In their approach, they associate regions with high contribution to a model's classification decision derived from saliency maps to image segments that represent contextual information i.e., key features of the displayed objects. Similarly, we introduce an explanation framework for tree species classification that links image regions with high Finer-CAM saliency to segments representing structural tree features in 2D side-view images of TLS 3D point clouds. This analysis can further enhance the understanding of how structural tree features (e.g., crown, stem) contribute to the classification decisions of these models.

In this study, we aim to contribute to enhancing interpretability of CNN-based tree species classification models utilizing TLS 3D point clouds. More specifically, we aim to better understand the classification decisions of a YOLOv8-based tree species classification approach when assigning a given 2D side-view image derived from a TLS 3D point cloud of an individual tree to a target species class. We pursue this goal through two main steps: 1) by the application of a novel framework designed to enhance the interpretability of YOLOv8's classification decisions linking Finer-CAM saliency map regions of high discriminative power to image segments representing structural tree features; and2), by evaluating the ability of Finer-CAM to correctly highlight features in 2D side-view images based on their contribution to YOLOv8`s classification decisions, through an analysis of its faithfulness. We further assume that model performance is influenced by visibility of internal tree structure (e.g., branches) in 2D side-view images derived from TLS 3D point clouds. To gain insights into how internal tree structure visibility influences model performance, we conducted additional experiments in which we systematically perturbed 2D side-view images.

## Methods

### *Single tree TLS point clouds*

We used a subset of the FOR-Species20k benchmark data set (Puliti et al. 2025) which consists of proximal laser scans (UAV-LS, MLS, TLS) of 19 962 individual trees containing 33 tree species from Europe, Canada, Australia, and New Zealand. FOR-Species20k is a compilation of earlier existing data sets. It shows strong class imbalance, variable tree height distributions among tree species classes, and an unbalanced representation of laser scanning platforms (UAV-LS, MLS, TLS), and data sources (research labs providing the data). To enhance the generalisability of the results by increasing the variability of data sources available per target class, a subset of the data set was created. This subset included only TLS data of tree species for which data is provided by a minimum of three different data sources. This filtering resulted in a reduction of the dataset to seven tree species, which are Silver birch (*Betula pendula*, Roth), Eruopean beech (*Fagus sylvatica*, Linnaeus), Eruopean Ash (*Fraxinus excelsior*, Linnaeus), Scots pine (*Pinus sylvestris*, Linnaeus), Norway spruce (*Picea abies*, Linnaeus), Douglas-fir (*Pseudotsuga menzisii*, Franco), and English oak (*Quercus robur*, Linnaeus). To compensate for the pronounced class imbalance, we reduced the number of samples for tree species classes with more than 1000 instances by randomly selecting 1000 trees per class. We randomly split the TLS point clouds for every species into training (90 %) and test data (10 %). We employed a cross-validation approach, where the training data was randomly shuffled and divided into five folds, with each fold serving once as the validation set during training and hyperparameter tuning. The test data was reserved exclusively for evaluating the performance of the final trained models. Figure 1 shows specifications of the training/validation and the test sets in more detail.

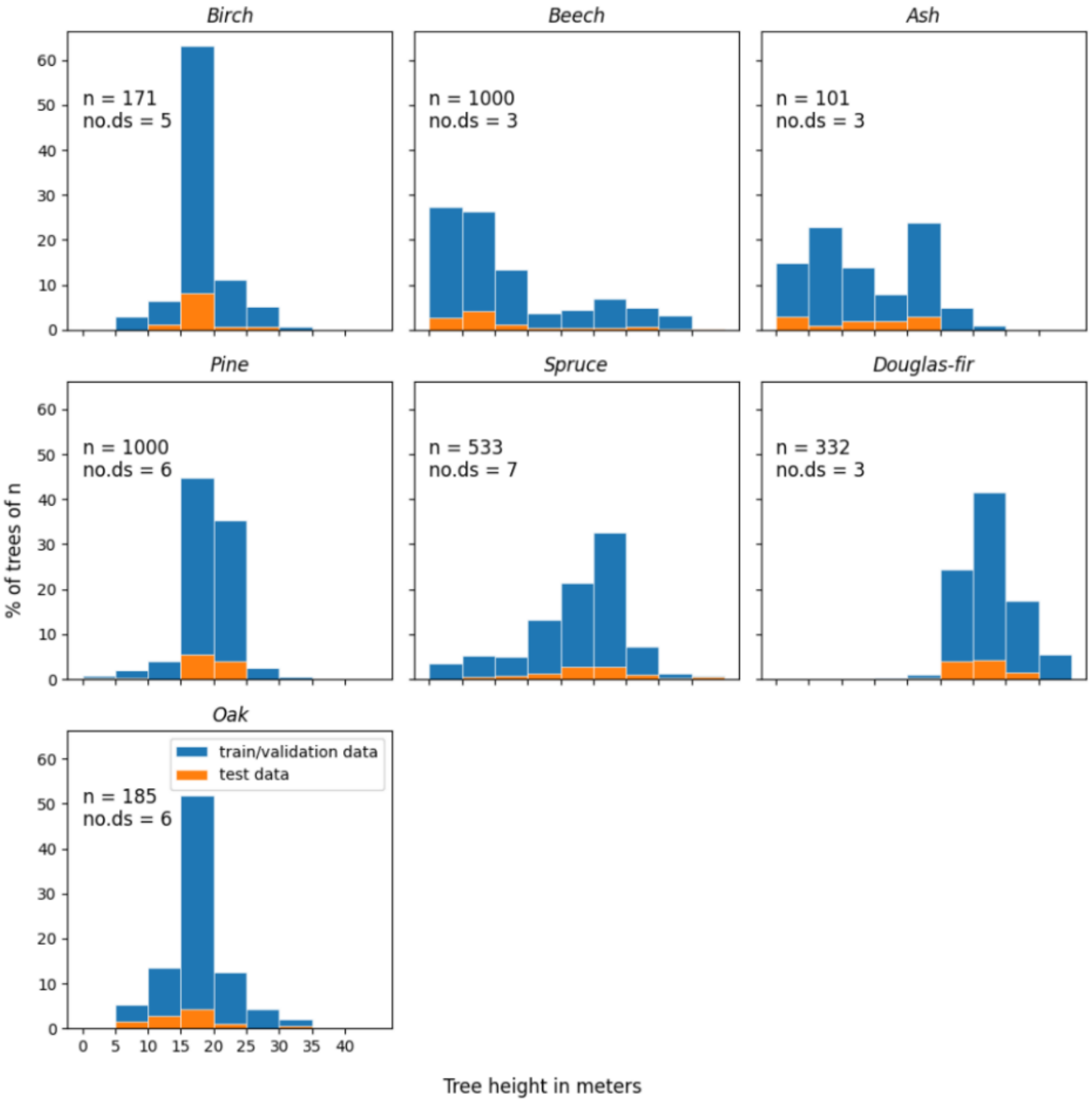

Figure 1 Tree species-wise tree height distributions in the data set splits (training/validation, test) for 2445 individual tree TLS 3D point clouds from the FOR-Species20k data set (Puliti et al. 2025) used in this study. n indicates the numbers of trees per tree species. No.ds indicates the number of data sources.

### *Orthographic 2D projections of TLS point clouds*

Similar to Zou et al. (2017) and Seidel et al. (2020) we used an orthographic projection approach to generate 2D side-view projections of TLS point clouds of the training and test data sets (see Figure 2 a.- b.). In the following, “2D side-view images” refers to 2D side-view projections derived from TLS 3D point clouds. We followed a similar approach for the projection as presented in Puliti et al. (2025).

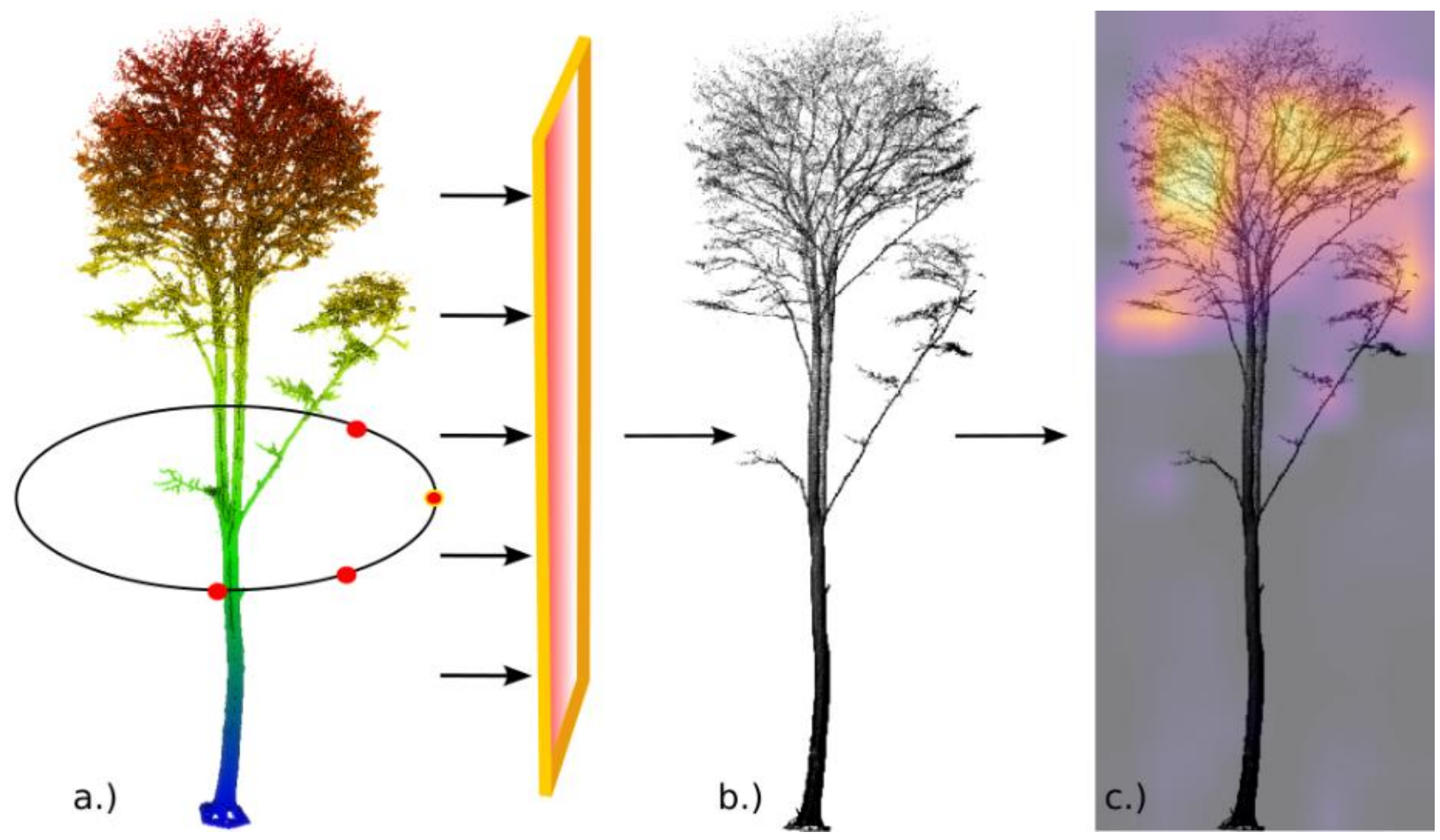

Figure 2 Workflow of the orthographic 2D projection of individual tree point clouds and saliency map generation. a.) Red circles indicate the positions from which the points are projected onto 2D planes. b.) From each position the points are orthographically projected onto a 2D raster to create a 2D side-view image. c.) Finer-CAM saliency map projected onto a side-view image.

The points of some laser scans in the data set form regular point patterns, due to scanner properties. These regular point patterns, when associated with tree species, may hinder generalizability of the classification model and could lead to shortcut learning since it may learn to differentiate tree species based on point distribution patterns. Consequently, we shifted all points by up to half a centimetre in random directions to confound regular point patterns.

To create the 2D side-view images we projected all TLS points to a 2D plane which was rotated around the *z-axis* (which represents elevation) by angles ∝ of 0°, 45°, 90°, and 135° in reference to north. Each plane is represented by two vectors u and v, where v points up (origin to tree top) and u spans the second direction of the plane. Computing the dot product of a given TLS point coordinate p with u and v yields the x and y coordinates within the plane:

$$u = [cos\ (\propto)\ , sin\ (\propto)\ , 0] \tag{1}$$

$$v = [0\ ,0\ , 1] \tag{2}$$

$$x = p\ \cdot\ u \tag{3}$$

$$y = p\ \cdot\ v \tag{4}$$

where ∝ is the rotation angle and p is a TLS point. After each projection we created a raster with a pixel size adjusted to the point density of the point cloud. This step ensures that every tree point cloud is projected over the entire image canvas. We calculated the point density adjusted pixel size in meters ($\Delta px$) as the theoretical space occupied by each point in the xy plane using the following equation:

$$\Delta px\ =\ \sqrt{(w\ \times h)\ \div\ n} \tag{5}$$

where h is the height of the tree on the z-axis of the point cloud, w is the maximum width of the tree on the y-axis of the point cloud and n is the number of points in the point cloud. We then aggregated the points along the x-axis of the point cloud by counting all points within each cell of the raster and stored the point count as pixel values.

To mitigate the fact that extreme pixel values dominate the final grayscale side-view images, we log-transformed the pixel values. Finally, we stored each raster as 640 pixels (px) × 640 px grayscale image.

### *YOLOv8 for tree species classification*

Our approach builds on the tree species classification method YOLOv5 presented in Puliti et al. (2025). While the original method used YOLOv5, we used YOLOv8 (Jocher et al. 2023) here as it demonstrates superior performance compared to YOLOv5 on the COCO benchmark data set (Lin et al. 2014) and supports larger image sizes beyond 124 px × 124 px. This enhancement allows for higher spatial resolution in 2D side-view images, improving the interpretability of classification results by enabling visualization of tree features with more details. YOLOv8 is characterised by a backbone and a classification head (Jocher et al. 2023). The backbone starts with two consecutive convolutional modules followed by four c2f blocks, of which the first three c2f blocks are each followed by a convolutional module. Each of the c2f blocks consists of a convolutional module that is followed by concatenated bottleneck modules (e.g., Heravi et al. 2018) that are followed by a convolutional module. In the first and last, and second and third c2f blocks three and six bottleneck modules are used, respectively. All convolutional modules in the backbone have a padding of 1, a stride of 2, and a kernel size of 3. The convolutional modules in the c2f modules have a padding of 0, a stride of 1 and a kernel size of 1. In all convolutional modules batch normalisation (Ioffe & Szegedy 2015) is performed and the SILU activation function (Elfwing et al. 2018) is applied. In the classification head a convolutional module is followed by a 2D adaptive average pooling layer, a dropout function (Hinton et al. 2012) and a linear transformation function.

The performance of a CNN is influenced by the training data it receives. Even minor changes in the training set or shifts in data distribution can result in unstable model predictions (Riley and Collins 2023). To mitigate this issue and enhance the reliability of our analysis, we evaluated the outputs of five independently trained YOLOv8 models. These models were trained using a five-fold cross-validation strategy, with each model utilizing different data subsets for training and validation. Additionally, we leveraged pre-trained weights (yolov8s-cls.pt, Jocher et al. 2023) originally trained on the COCO data set. During training we used a learning rate scheduler which is based on the “1cycle policy” described by Smith and Topin (2019) with an initial learning rate of $10^{-2}$ and a minimum learning rate of $10^{-4}$. We reduced the initial learning rate during each test in steps of one tenth of the previously tested initial learning rate starting from $1 \times 10^{-1}$ until a lower threshold of $1 \times 10^{-4}$ was reached.

Based on these tests we selected an initial learning rate of $1 \times 10^{-2}$ for the training of all models. We further set the batch size to 32 since it was the largest batch size suitable for the used hardware (GPU NVIDIA GeForce RTX 3080). We employed a stochastic gradient descent optimizer with weighted cross-entropy loss to compensate for class imbalance. We trained each model for 50 epochs and selected the weights that produced the highest overall accuracy on the validation fold without overfitting for further analysis.

For performance assessment we applied the trained models to the test data set in which each tree is represented by four 2D side-view images. To compute the final tree-wise species predictions we selected the class label with the highest average output logits among the class logits cast by a model of each tree’s four 2D side-view images.

We then calculated model- and class-wise F1-scores, recall and precision, and computed confusion matrices. To assess the overall performance of all models we calculated the macro averages of the model-wise F1-scores, recall and precision.

### *Finer-CAM saliency calculation*

Finer-CAM (Zhang et al. 2025) is a feature attribution method tailored for CNNs trained on image data and building upon Grad-CAM (Selvaraju et al. 2020). Its goal is to provide a value of *saliency*, usually normalized to the 0-1 range, to each of the input pixels of the image, according to its *importance* within determining the prediction of the model. It is an inherently *local* method, i.e., the saliency scores are only applicable to the single input being analysed, and do not generalize to global rules.

Finer-CAM generates the saliency map by fixing an intermediate layer $l$ and storing the information provided by activations and gradients, as explained next. Additionally, the definition of two classes the model is trained to recognize is required: (i) the category $c$, for which we want to generate the saliency scores, and (ii) a *contrastive category* $c'$ (i.e., the most similar class), which acts as a *counterfactual* to the class $c$. In this sense, the Finer-CAM output will highlight pixels with positive attribution towards class $c$, but **not** towards class $c\,'$.
The gradients are calculated on the difference between the output logits $\mathrm{f}(x)_c$ and $\mathrm{f}(x)_{c'}$ and are computed with respect to the activations of the layer $l$. We formally indicate this as $\frac{\partial\left(\mathrm{f}(x)_c - \gamma \cdot \mathrm{f}(x)_{c'}\right)}{\partial A^{(l)}}(x)_c$, with $A^{(l)}$ indicating the activations of layer $l$ and $\gamma$ is a comparison strength coefficient. The gradients are averaged on the spatial dimensions $h, w$, to produce a vector $\alpha$. For a generic channel $k$, being

$$\alpha_k = \frac{1}{Z} \sum\nolimits_{h,w} \frac{\partial(\mathrm{f}(x)_c - \gamma \, \cdot \, \mathrm{f}(x)_{c'})}{\partial {A^{(l)}}_{k,h,w}}, \tag{6}$$

where $Z$ is the product of the spatial dimensions of the activation map. The vector $\alpha$ highlights the saliency, according to the gradients, of each of the channels within the map. It is further combined with the activations. In that way saliency $s_k$ is computed by

$$s_k = \alpha_k \cdot {A^{(l)}}_k, \tag{7}$$

where ${A^{(l)}}_k$ indicates the matrix of the activation map for channel $k$.

Finally, using the $ReLU$ function, Finer-CAM suppresses negative values, retaining only *positive attributions*, i.e., indicating only those pixels who contribute positively towards class $c$ (and not $c'$):

$$FinerCAM_k = ReLU(s_k). \tag{8}$$

Additionally, since usually the spatial dimensions of the activation map of layer $l$ are smaller than the ones of the image, upsampling via bilinear interpolation is applied to the output of Finer-CAM to match the spatial dimension of the image, thus allowing the saliency map to be overlaid on top of the image (see Figure 3).

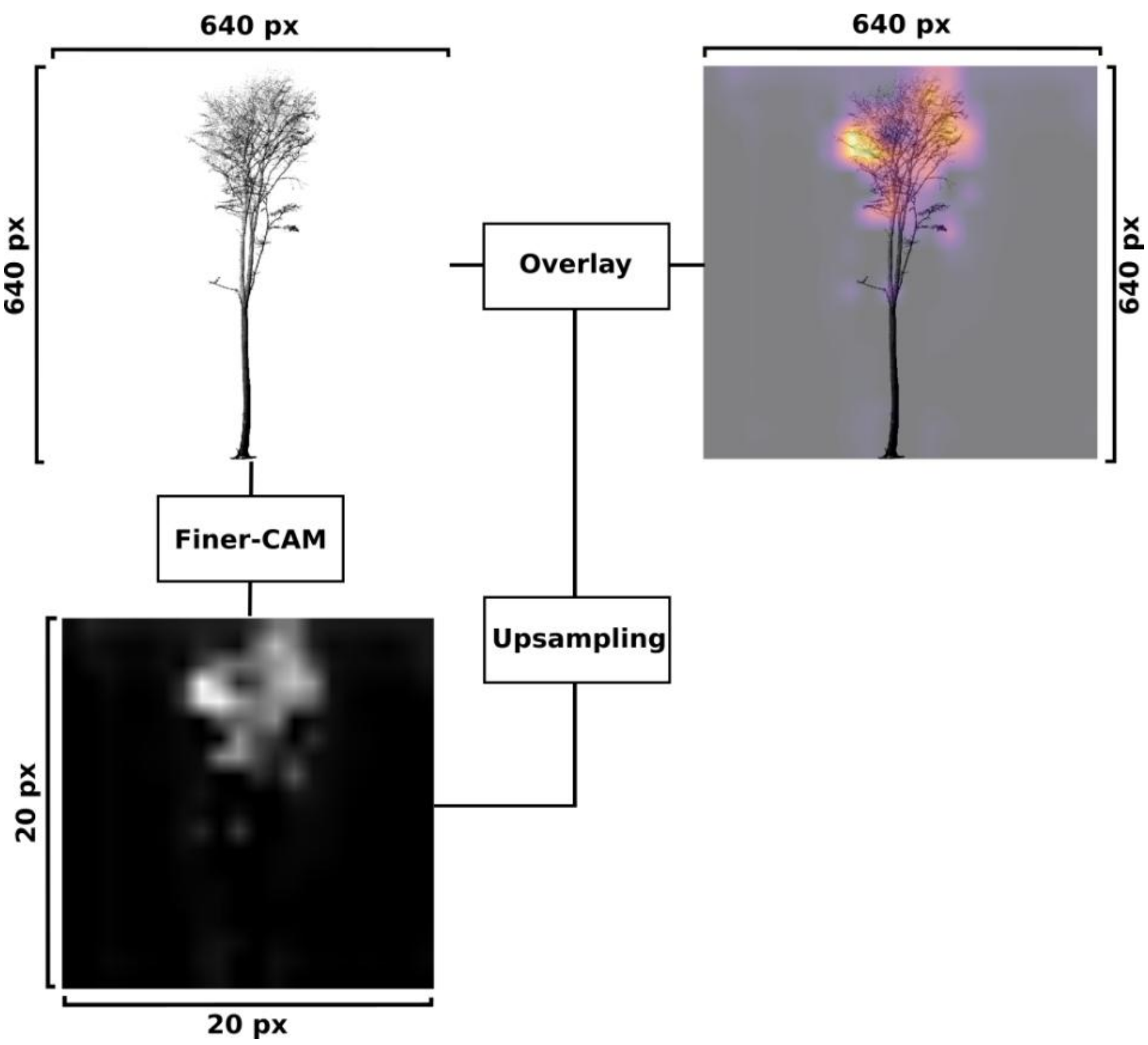

Figure 3 Upsampling of the output of Finer-CAM to the spatial dimension of the 2D side-view images and overlay of the upsampled saliency map on to a 2D side-view image.

### *Saliency map generation*

In Finer-CAM saliency maps image features are highlighted that contribute to the classification of a target class but not to contrastive classes most similar to the target class. Consequently, saliency maps are model and input specific. Feature contribution to the classification decisions of a model can vary among images of the same target class and different features can contribute to a given target class for different models. To compensate for these effects, we comprehensively analyse saliency maps of all five trained YOLOv8 models for the same selection of 2D side-view images of the test data set.

For the systematic analysis of saliency maps of all models we randomly selected 20 trees from the test data of each of *Beech, Pine, Spruce,* and *Douglas-fir*. The inclusion of more trees in our analysis would have increased evaluation time substantially and was hence dismissed. Since the test data set includes less than 20 trees of *Birch, Ash,* and *Oak* we conducted no further sub-sampling. From each of the selected trees we randomly selected one of the four 2D side-view images when it was correctly classified by all five models. Consequently, we selected 20 2D side-view images of *Beech*, *Pine*, Spruce *Douglas-fir*, 18 2D side-view images of *Birch*, 19 2D side-view images of *Oak*, and 9 2D side-view images of *Ash* for further analysis.

For each of the five YOLOv8 models we generated Finer-CAM saliency maps for each of the selected 2D side-view images using the Python library pytorch-grad-cam by Gildenblat (2021). We selected the last so-called c2f block in each model, which is the last layer before the classification layer, as the target layer for the saliency map generation. Using the last network layer for the generation of saliency maps is frequently applied (Selvaraju et al. 2017, Grad-CAM++ (Chattopadhay et al. 2018), Muhammad et al. 2020, Zhang et al. 2025)).

Finer-CAM allows for the aggregation of activation weights of multiple contrastive classes to produce more comprehensive saliency maps for a target class. We aggregated three contrastive classes most similar to a target class i.e., classes of which the output logits rank second, third and fourth after the output logit of the target class. Similar to the procedure conducted by Zhang et al. (2025) who show that this produces overall the best results. Further we set the comparison strength coefficient $\gamma$ to 1. The comparison strength coefficient ranges between 0 to 1 where lower values produce finer and higher values produce coarser saliency maps, respectively.

### *Heuristic integration of image partitioning and saliency mapping*

To link recognizable tree features characterising structural features of a tree in 2D side-view images with the predictions of the five models—and to reduce potential cognitive biases in saliency map interpretation (e.g., Mohseni et al. 2021; Bertrand et al. 2022)—we developed a heuristic approach to classify pixels from the saliency maps as related to tree, tree stem or tree crown. We further subdivided the tree crown into four sub-segments as shown in Figure 4.

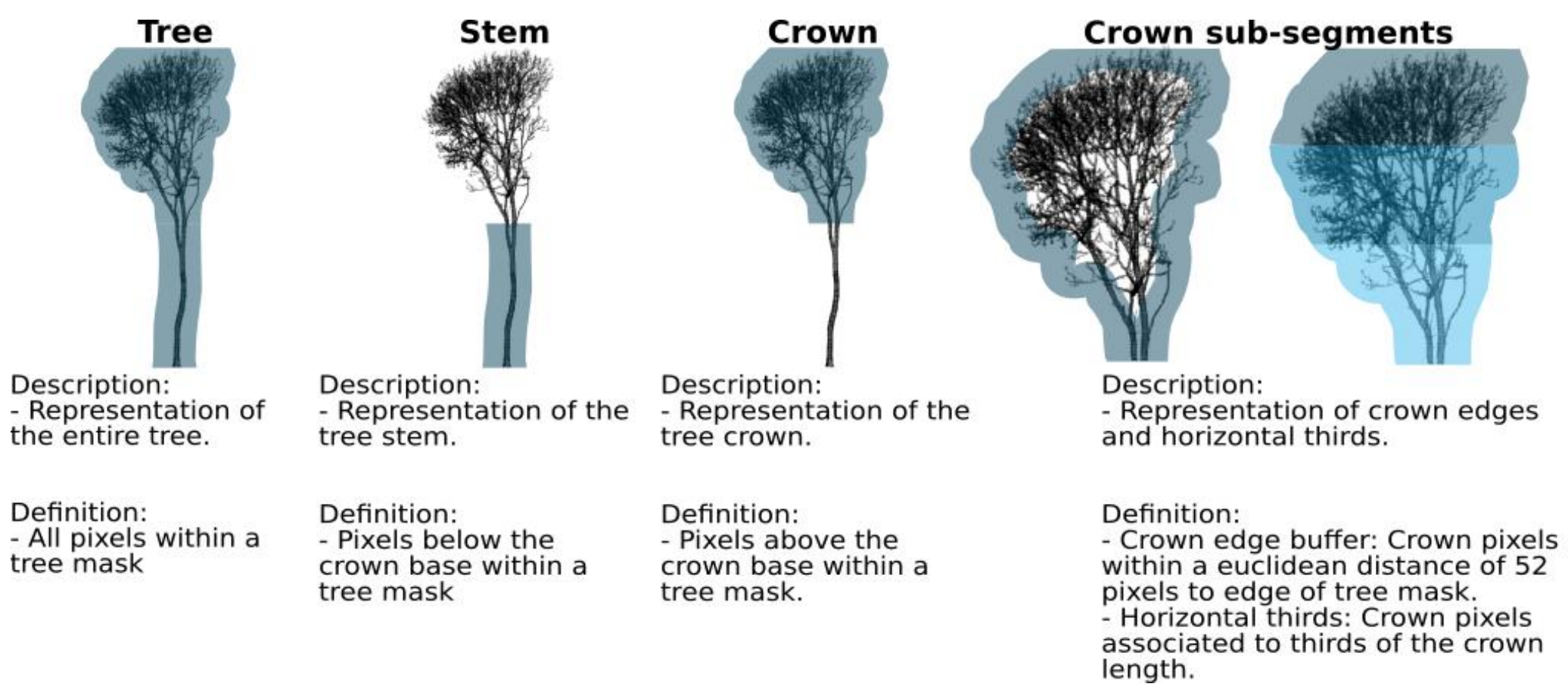

Figure 4 Descriptions and definitions of the image segments associated with structural tree features to which we attributed salient pixels of Finer-CAM saliency maps.

### *Image partitioning*

To partition individual trees in the 2D side-view images, we generated binary tree masks. In a first step we calculated tree contours by the application of a Gaussian blur filter (kernel size: 15 × 15 pixels) in which $\sigma$ is computed from the kernel size as follows:

$$\sigma = 0.3 \cdot \big((k-1) \cdot 0.5 - 1\big) + 0.8 \tag{9}$$

where $k$ is the one-dimensional kernel size in pixels. In a next step we applied binarization by setting pixel values to 255 and 0 when pixel values in the blurred images are higher or lower than a threshold of 0.0001, respectively. Finally, we extracted tree contours using the contour detection algorithm presented by Suzuki and Abe (1985).

The upsampling via interpolation is conducted with a scaling-factor of 32 during the generation of the saliency maps. This leads to a reduction in localization accuracy of the salient regions in the upscaled maps (see Figure 3).

To mitigate the effects of the localization errors we created a buffer with a distance of 32 px around the tree contours for the generation of the tree masks. We consider pixels within the buffer as tree pixels.

To further partition into crown and stem regions, the height of the crown base was manually identified on each tree mask as the point where the stem transitions into multiple crown branches. In a straightforward scenario, pixels above the crown base were assigned to the crown, while those below were attributed to the stem. However, in instances where branches extended below the crown base, this method proved insufficient. To resolve this, the lowest crown pixel was also manually marked.

A path connecting the topmost and the lowest crown pixels was then computed using the *route_through_array* function from the scikit-image library (van der Walt et al. 2014). This algorithm determines the path of 1 pixel width that minimizes the cost between two points, where the cost of 1 is assigned to pixels within a tree, and the cost of 2 to pixels that fall outside of the tree. Prior to path detection, isolated pixel groups smaller than 64 pixels were removed to prevent noise interference. The resulting path approximates the stem's longitudinal axis. A buffer was then generated around this path, extending from the crown base to the lowest crown pixel, with the buffer width defined as the pixel width of a representative horizontal stem section (determined on a case-by-case basis). Tree pixels within this buffer and those below the lowest crown pixel were classified as the stem, while all remaining tree pixels were designated as the crown.

Recognizing the importance of crown morphology for species differentiation, the crown segment was further subdivided. The edges of the crown, which capture small branches, species-specific branching patterns, and terminal branch segments, were delineated by including all crown pixels within a Euclidean distance of 52 pixels from the edges of the tree masks. Moreover, inherent occlusions in TLS-derived point clouds often lead to under-representation of the crown tops, resulting in a negative correlation between tree height and point density. To evaluate the potential impact of these effects on species classification using YOLOv8, the crown was horizontally divided into three segments (base, middle, top), each corresponding to one-third of the crown's total height. Salient pixels were then assigned to their respective horizontal segments. Descriptions and definitions of the image segments are presented in Figure 4.

### *Evaluation of saliency maps*

To analyse which tree species the models regard as similar to each other we tracked the three contrastive tree species —i.e. the species of which the models' output logits rank second, third and fourth after the output logit of the target species— of each saliency map. We then calculated the relative frequency of each contrastive species per target tree species over all saliency maps and models.

We quantitatively assessed the importance of predefined tree segments (i.e., tree, crown, stem, and crown sub-segments) to the models predictions by attributing salient pixels to these segments.

Since we are only interested in those regions of the 2D side-view images that have the highest contributions to class predictions, we counted pixels in the saliency maps as salient pixels when the pixel value was above a certain threshold which we computed per saliency map using Otsu's method (Otsu 1979).

To ensure comparability between saliency maps and tree species we calculated the ratio of the number of salient pixels attributed to a segment and the number of salient pixels attributed to the segment one level above in the structural hierarchy. In Table 1, we define these ratios of salient pixels.

Table 1. Calculated ratios of the number of salient pixels (defined as pixels with a pixel value > a threshold computed by the application of Otsu's method) in each segment and abbreviations used in the following to refer to these.

| Segment | Definitions of No. of salient pixels | Ratio of salient pixels |
|---|---|---|
| Tree | Nst = No. of salient pixels in tree segment<br>Nstot = No. of salient pixels in saliency map | $rNst = \frac{Nst}{Nstot}$ |
| Stem | Nss = No. of salient pixels in stem segment | $rNss = \frac{Nss}{Nst}$ |
| Crown | Nsc = No. of salient pixels in crown segment | $rNsc = \frac{Nsc}{Nst}$ |
| Crown base | Nscb = No. of salient pixels in crown base segment | $rNscb = \frac{Nscb}{Nsc}$ |
| Crown middle | Nscm = No. of salient pixels in crown middle segment | $rNscm = \frac{Nscm}{Nsc}$ |
| Crown top | Nsct = No. of salient pixels in crown top segment | $rNsct = \frac{Nsct}{Nsc}$ |
| Crown edge buffer | Nsce = No. of salient pixels in crown edge buffer segment | $nNsce = \frac{Nsce}{Nsc}$ |

We further aggregated the ratios of salient pixels for all segments at tree species level and over all tree species by calculating the arithmetic means and the standard deviations. On tree species level we further generated box plots of the ratios of salient pixels for each segment. To evaluate differences among tree species, we conducted statistical significance tests on the ratios of salient pixels of all segments. Normality of the data was assessed using the Shapiro-Wilk test, which revealed that the ratios of salient pixels in some segments were not normally distributed. Consequently, we employed the Dunn's test ($\alpha = 0.05$) for pairwise comparisons of the tree species. To mitigate the errors due to the performance of multiple significance tests (see Bender & Lange 2001) we used adjusted p-values applying the Holm-Bonferroni method (Holm 1979).

### *Image perturbation experiments*

The appearance of trees in the 2D side-view images is influenced by both structural tree properties — such as habitus and internal tree structure (i.e., branching patterns)— and by technical features of point clouds. Structural tree properties are directly observable in the 2D side-view images, whereas technical features of point clouds like point densities, affect pixel values of the 2D side-view images. These data features are potentially available to the models for tree species differentiation. To investigate the sensitivity of YOLOv8 towards certain features we conducted experiments in which we assessed model performance when features are impaired or emphasised in the 2D side-view images. We created eight additional data sets in each of which we perturbed the 2D side-view images of the data set. In all experiments we used the same data set split, training procedure and performance assessment presented in sections *Single tree TLS point clouds* and *YOLOv8 for tree species classification*, respectively. Details of each experiment (Exp) are presented below.

#### *Experiment-shape*

We define tree shape as the area an individual TLS 3D point cloud side-view projection creates when all empty spaces within the tree crown are filled. With this definition we assume that tree shape relates to a tree's habitus. To investigate the capabilities of YOLOv8 to differentiate tree species solely on tree shape we transformed the 2D side-view images into binary tree shape images that reduce the structural complexity and only display the tree shape (see Figure 5). For the creation of these tree shape images we applied a dilation operation from the Python library OpenCV (Bradski 2000) to the 2D side-view images. We applied a moving window of circular shape with a radius of 15 px and asserted the maximum pixel value within the window to the centre pixel. Afterwards we tracked the outline contours of the dilated tree representation by the application of the *findContours* function of OpenCV (Bradski 2000) and set all pixels within the tree contours to black.

#### *Experiment internal tree structure visibility reduction*

To better understand the capabilities of YOLOv8 to leverage information on internal tree structure for species classification, we transformed the 2D side-view images which are gray-scale images into binary images, by setting every non-white (tree) pixel to black (see Figure 5). This transformation emphasizes the visibility of internal tree structure in the 2D side-view images, as it reduces information on point density that translates into the pixel values of the side-view images. To further analyse the sensitivity of YOLOV8 towards the visibility of internal tree structure we gradually reduced the quality of the binary 2D side-view images (see Exp-binary) by consecutively down- and upscaling them (see Figure 5). Since YOLOv8 allows only for the use of minimum image sizes of 32 px x 32 px this approach enables down scaling of the 2D side-view images below the minimum image size required. So, we were able to substantially reduce internal tree structure visibility in the 2D side-view images. We performed the resizing of the 2D side-view images using the Python library OpenCV (Bradski 2000). We down-scaled the 2D side-view images to sizes of 320 px, 160 px, 80 px, 40 px, 20 px, and 10 px by using a pixel area relation resampling approach, which allows for the reduction of the displayed level of detail while maintaining a low level of blurring. This method computes the target pixel value as the average of the pixel values in a corresponding area around the target point.

The area is defined as the factor of the ratios of the sizes of the x and y axes of the original 2D side-view image to the target sizes of the x and y axes. After down-scaling we applied a bilinear interpolation approach considering the four nearest pixels to the calculated target pixel position for up-scaling to 640 px x 640 px.

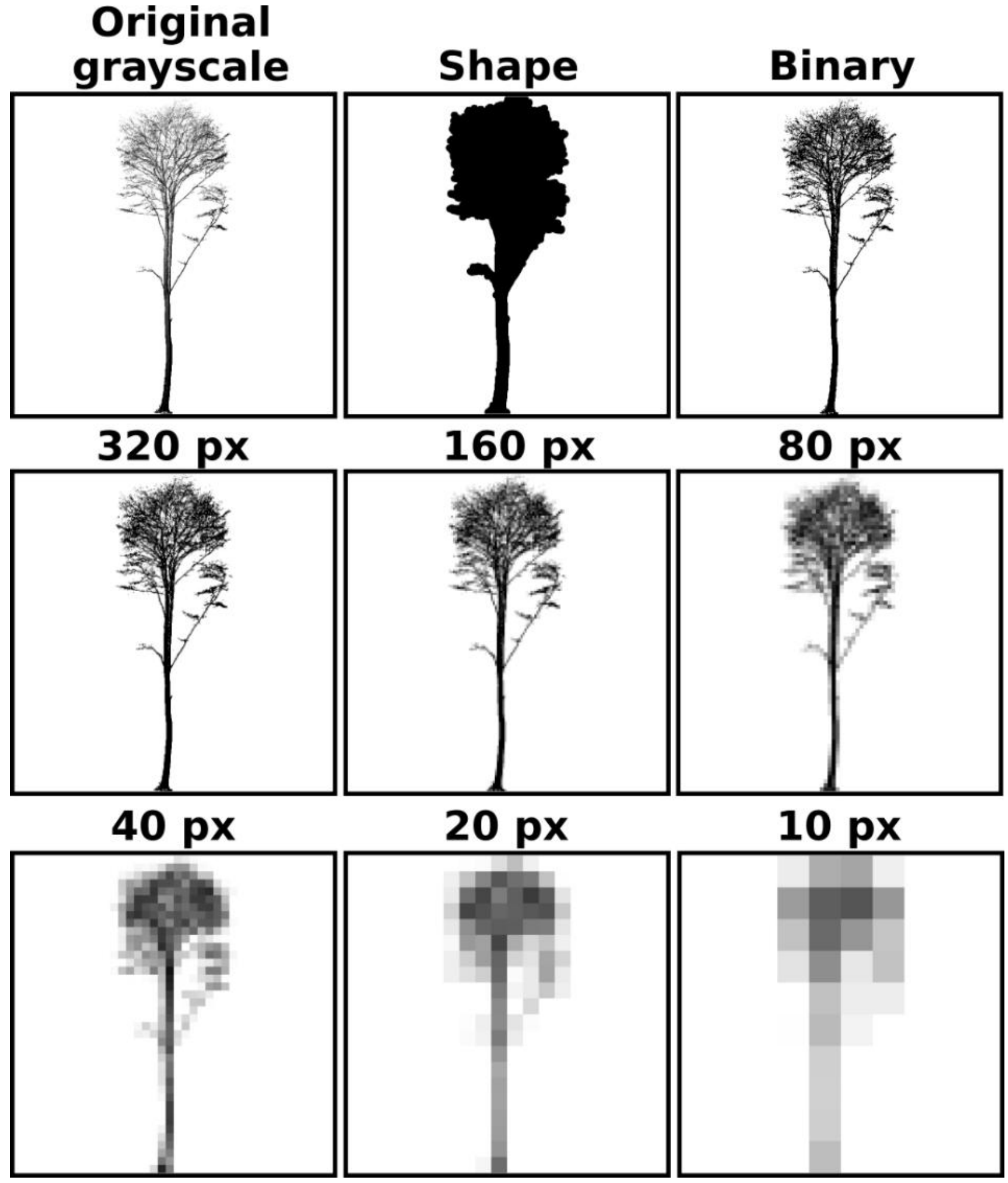

Figure 5 2D side-view image perturbation applied during the image perturbation experiments. We transformed the original grayscale 2D side-view images to binary grayscale images, to images in which internal tree structure is reduced by means of downsampling them to different image sizes (320 px, 160 px, 80 px, 40 px, 20 px, 10 px), and to images representing tree shape.

### *Analysing faithfulness of saliency maps*

For assessing the faithfulness of the Finer-CAM saliency maps we adopted a least relevant first (LeRF) pixel flipping feature removal strategy (Bach et al. 2015, Samek et al 2021), because of its simplicity while providing a sound assessment of faithfulness. Here we used the saliency maps of the five YOLOv8 models trained on the non-perturbed data set generated for the correctly classified 2D side-view images of the test data set. In a first step we ordered all pixels (tree and background pixels) of each 2D side-view images by saliency and perturbed pixels by setting them to white in a step-wise manner starting from lowest to highest saliency (see Figure 6). At each perturbation step we applied the YOLOv8 models trained on the non-perturbed data set to the 2D side-view images in which pixels were removed by flipping them to white and tracked the output logits (model confidence) of the target class (i.e. the true tree species). We then calculated the ratio between the model confidences at each pixel removal step and the model confidences when applied to the non-perturbed 2D side-view images.

Afterwards we calculated the average model confidence ratio at each step for each tree species and model. Finally, we aggregated the results over all models and generated curves depicting the average confidence of all models per tree species per perturbation step. For a LeRF removal strategy, the expectation is that the relative confidence remains overall stable with the first iterations of removal (since pixels with low importance are ablated); the confidence should drop in a monotonically increasing fashion as we approach the most important pixels. To ensure a better interpretability of the results we applied the same appraoch to random saliency maps, which we generated using the CAM approach presented by Zhou et al. (2015) but instead of using the original weights oft he models we used random weights.

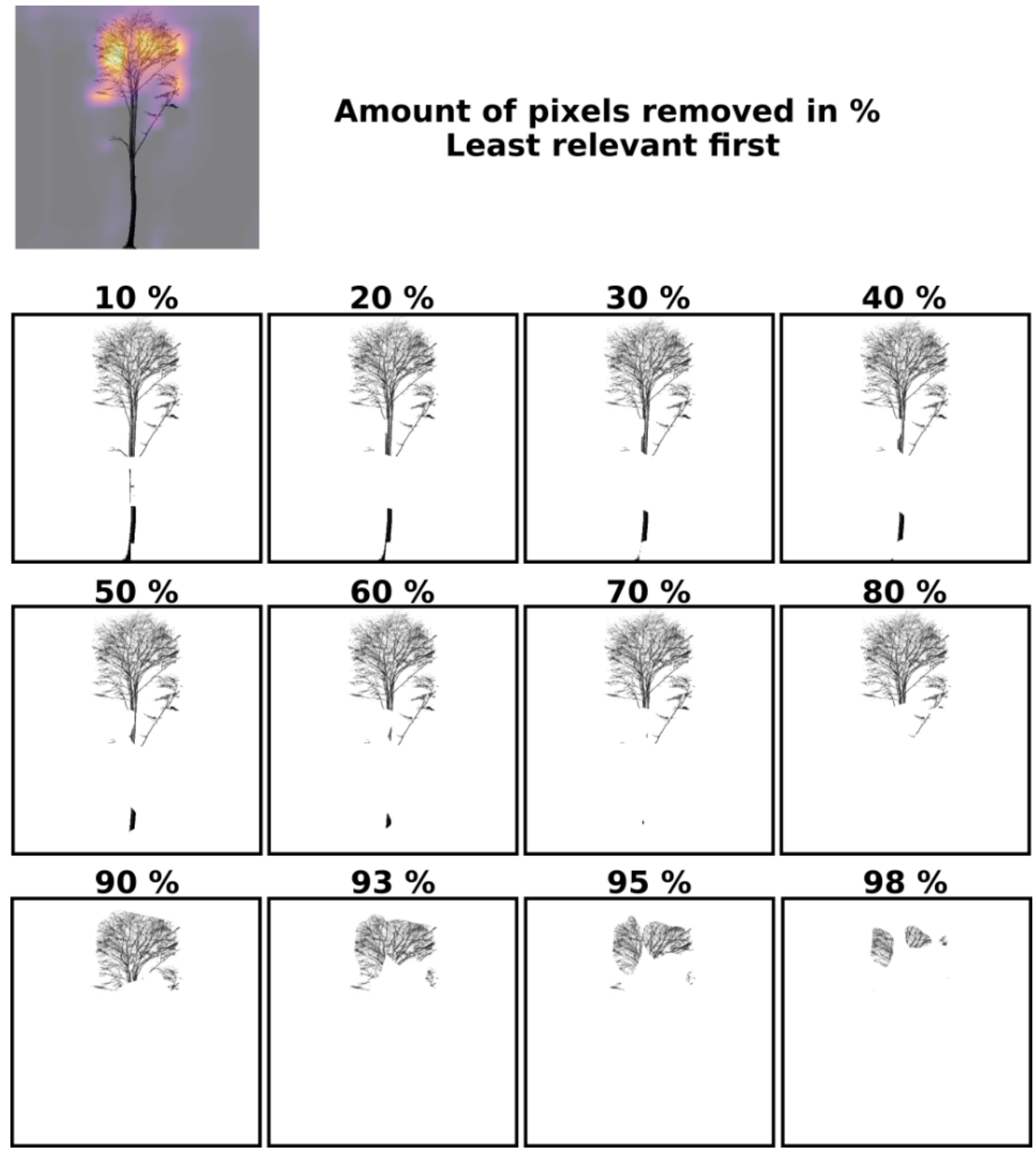

Figure 6 Example 2D side-view images of the same tree (Beech) showing the results of the application of a least relevant first pixel removal strategy based on Finer-CAM saliency maps (upper left corner) used to analyse the faithfulness of Finer-CAM.

## Results

In total we trained 45 YOLOv8 models for tree species classifications using 2D side-view projections of TLS 3D point clouds. For each trained YOLOv8 model, we selected the weights that produced the highest overall accuracy on the validation folds during training and applied the model to the test data.

### *Analysis of models applied to non-perturbed data set*

The following results are derived from the five YOLOv8 models trained on the non-perturbed training/validation data set using a five-fold cross validation approach, which we then applied to the non-perturbed test data set.

The highest overall accuracies during training were achieved after 40, 42, 44, 34, and 44 epochs for model 1, model 2, model 3, model 4, and model 5, respectively. On average the five trained models achieved an overall accuracy of 96 % (SD = 0.24 %), a macro F1-score of 92.5 % (SD = 0.19 %), a macro recall of 93.1 % (SD = 0.26 %), and a macro precision of 92.12 % (SD = 0.44 %) when applied on the test data set. A confusion matrix summarizing the performance results of all models is presented in Figure 7.

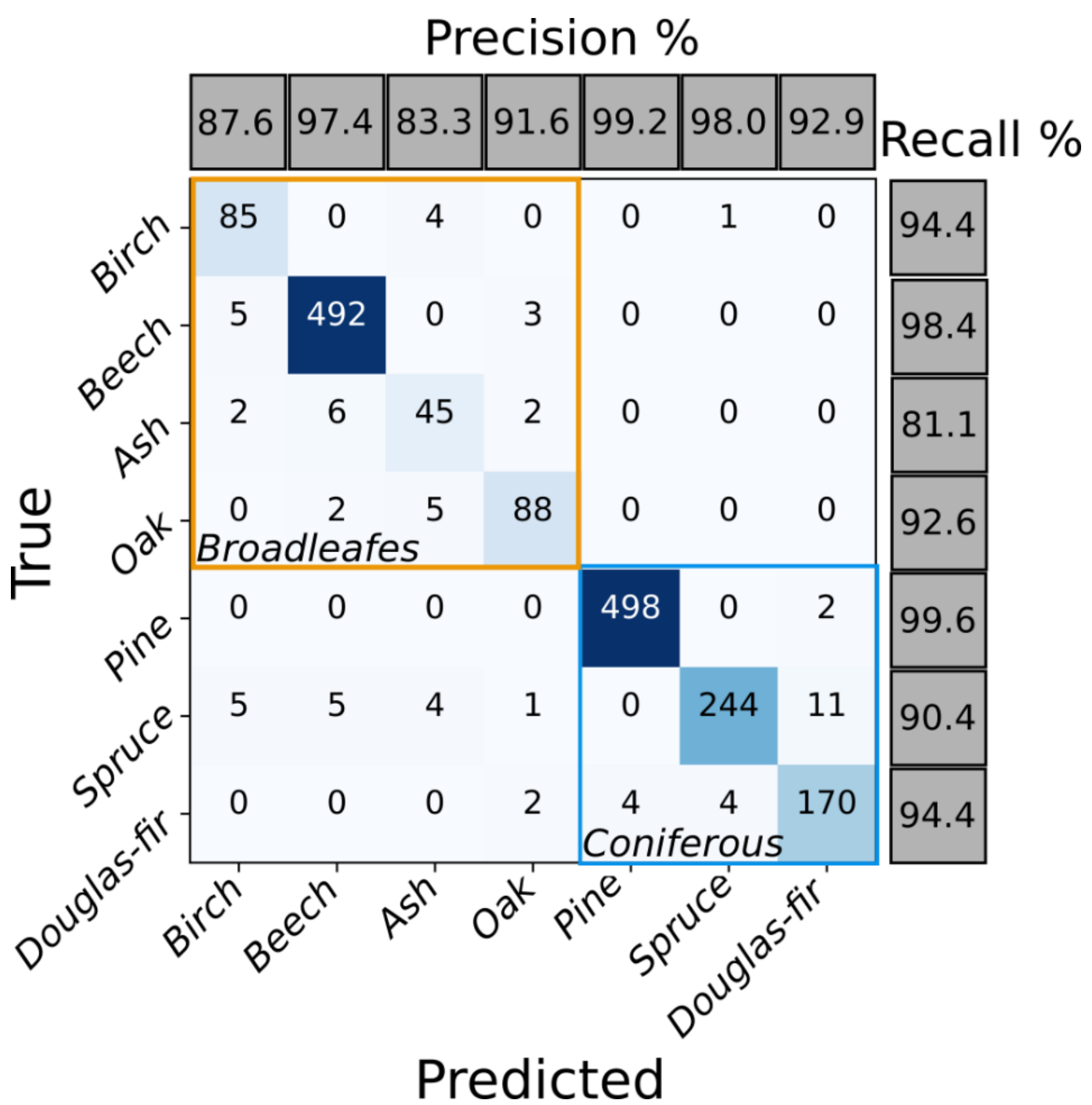

Figure 7 Confusion matrix, class-wise precision and recall summarizing the performance of the five YOLOv8 models trained on the non-perturbed training data set when applied to the non-perturbed test data set.

The application of Finer-CAM produced saliency maps in which the pixel values determine the importance of the pixel in regard to the prediction of a target tree species but are considered uncommon for three contrastive tree species most similar to the target species i.e., classes of which the output logits rank second, third and fourth after the output logit of the target class.

In Figure 8 we present the average ratios of the confidences (ratios of models output logits of target class) of the models to examine Finer-CAM saliency map faithfulness following a LeRF pixel removal strategy. Additionally, we compare these to the results of the application of a LeRF pixel removal strategy when pixels were removed based on random CAM saliency maps. In comparison the confidences of the models stay higher over all pixel removal steps when pixels are flipped based on their Finer-CAM saliency than according to a random saliency map. In addition, these results show that the confidences of the models remain steady until 80 % of less relevant pixels in regard to their Finer-CAM saliency are removed.

This indicates the capabilities of Finer-CAM to identify regions in the 2D side-view images of individual trees in relationship to their explanatory power to the classification decisions of the models rendering Finer-CAM faithful.

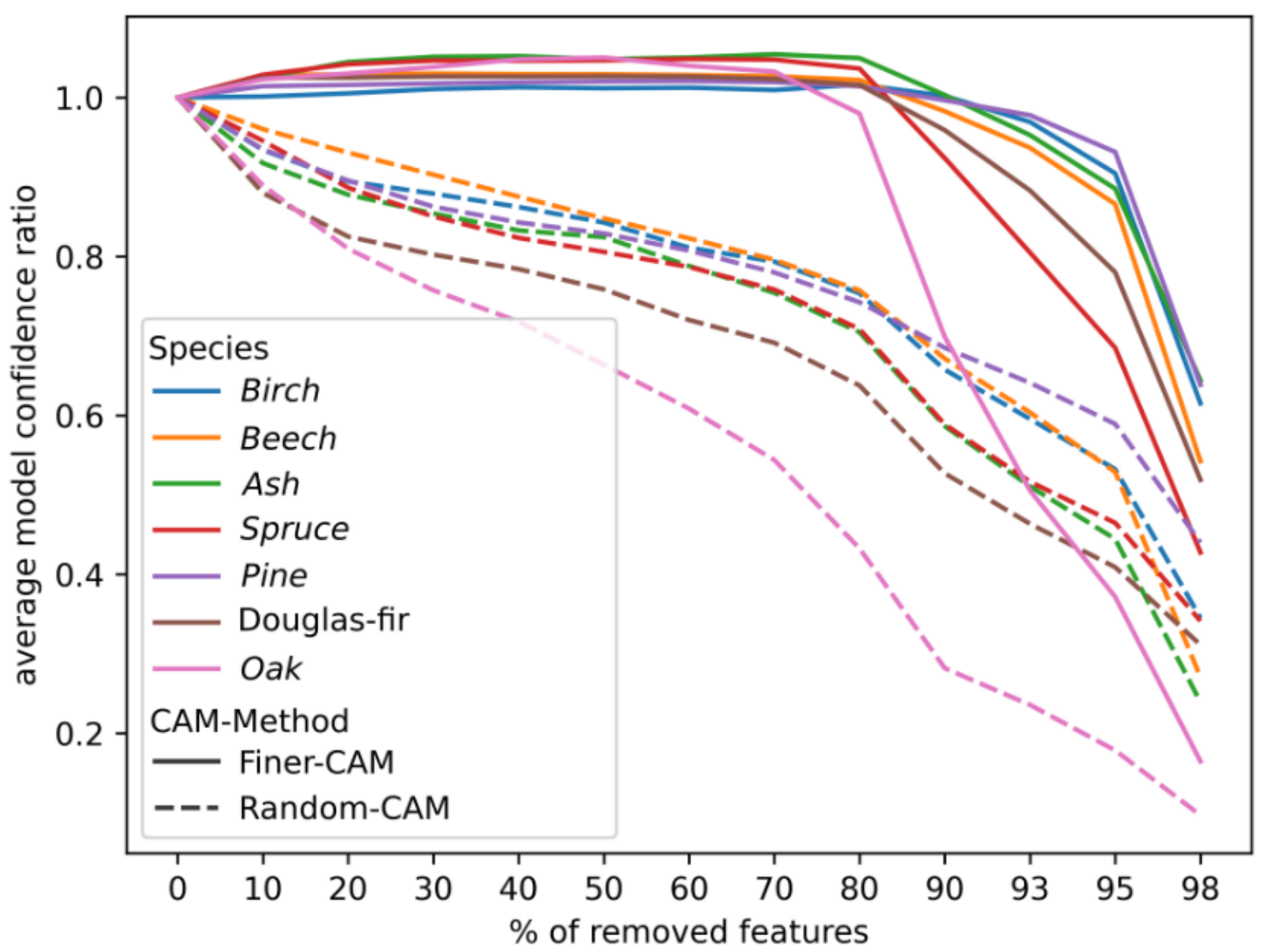

Figure 8 Average model confidence (output logits) ratio curves of target tree species when given percentages of pixels are removed (flipped) from 2D side-view images based on their saliency using a least relevant first approach. We used the YOLOv8 models trained on the non-perturbed data set. The filled lines depict model confidences when applied to 2D side-view images in which pixels are removed based on Finer-CAM saliency maps. The dashed lines depict model confidences when applied to 2D side-view images in which pixels are removed based on randomized saliency maps.

Example saliency maps of the YOLOv8 models trained and tested on the non-perturbed data sets generated for the same 2D side-view images of broad-leaved and coniferous tree species are presented in Figure 9 and Figure 10, respectively. These examples demonstrate that across all models similar image regions are deemed important in determining the model predictions by the application of Finer-CAM. This is least apparent for the example of *Spruce* for which image regions in the crown contribute most across all models, but the location of salient regions in the crowns differ across the models. Contrary, for the presented 2D side-view images of *Ash* all saliency maps across all five models highlight the same part of the stem, which represents a bend in the stem. On average across all models 64.4 % of the analysed saliency maps of *Ash* regions in the side-view projections depicting bends in stems are deemed most important to the model prediction by the application of Finer-CAM. In the examples of *Birch*, *Beech*, *Spruce*, *Pine*, *Douglas-fir*, and *Oak*, image regions representing branches are highlighted in the saliency maps. For *Pine* and *Douglas-fir* these branches are attached to the stems, whereas for the rest branches in the tree crowns are covered by salient regions. Additional examples of saliency maps of all tree species of model 1 are presented in Figure 11 and Figure 12.

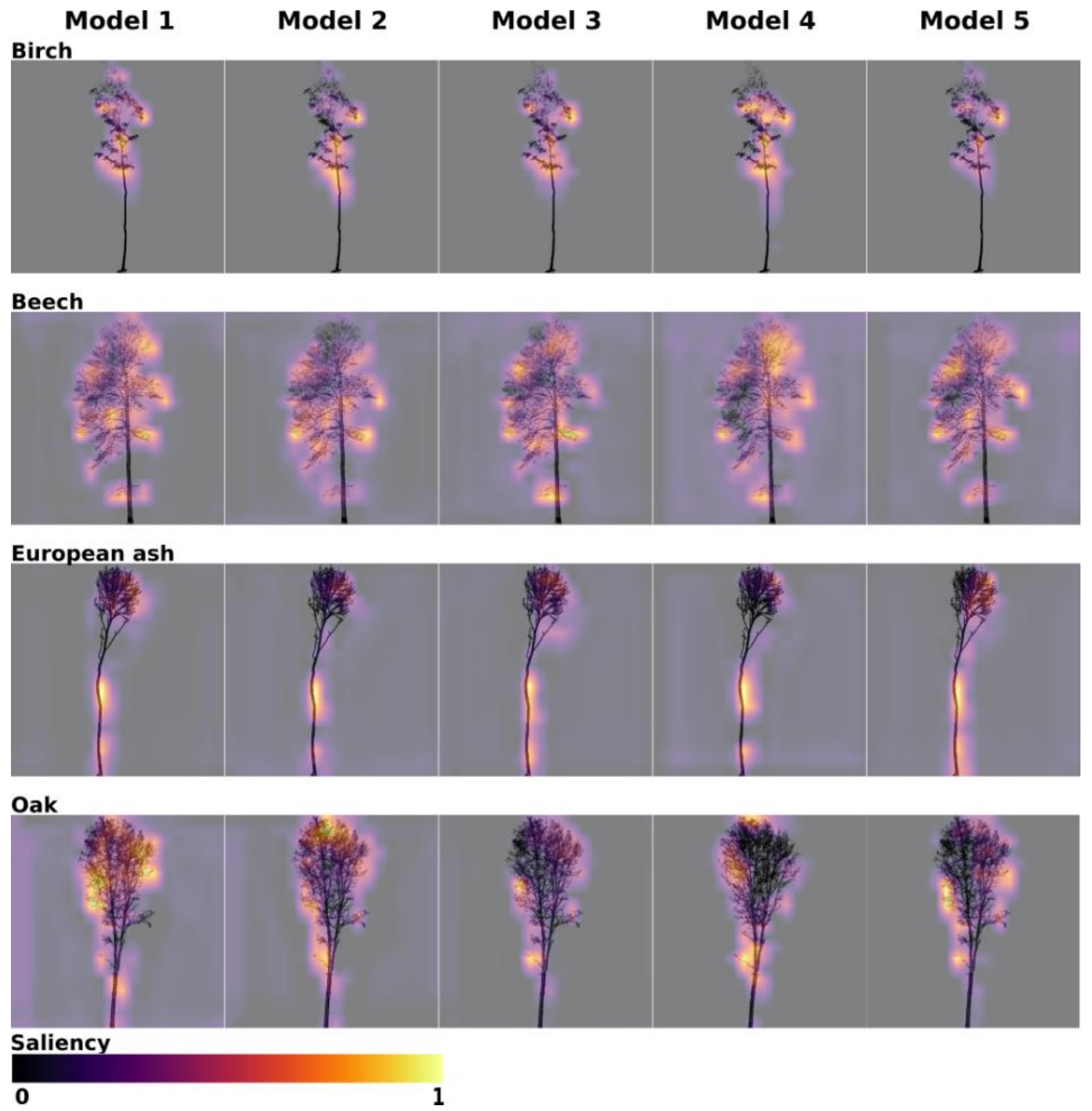

Figure 9 Example saliency maps of the YOLOv8 models trained on the non-perturbed data set of the same side-view projections for Birch, Beech, Ash, and Oak overlayed with the corresponding 2D side-view images. Layer activations for a species class are depicted in a colour gradient from purple to yellow, where yellow and purple correspond to the regions with high and low saliency, respectively.

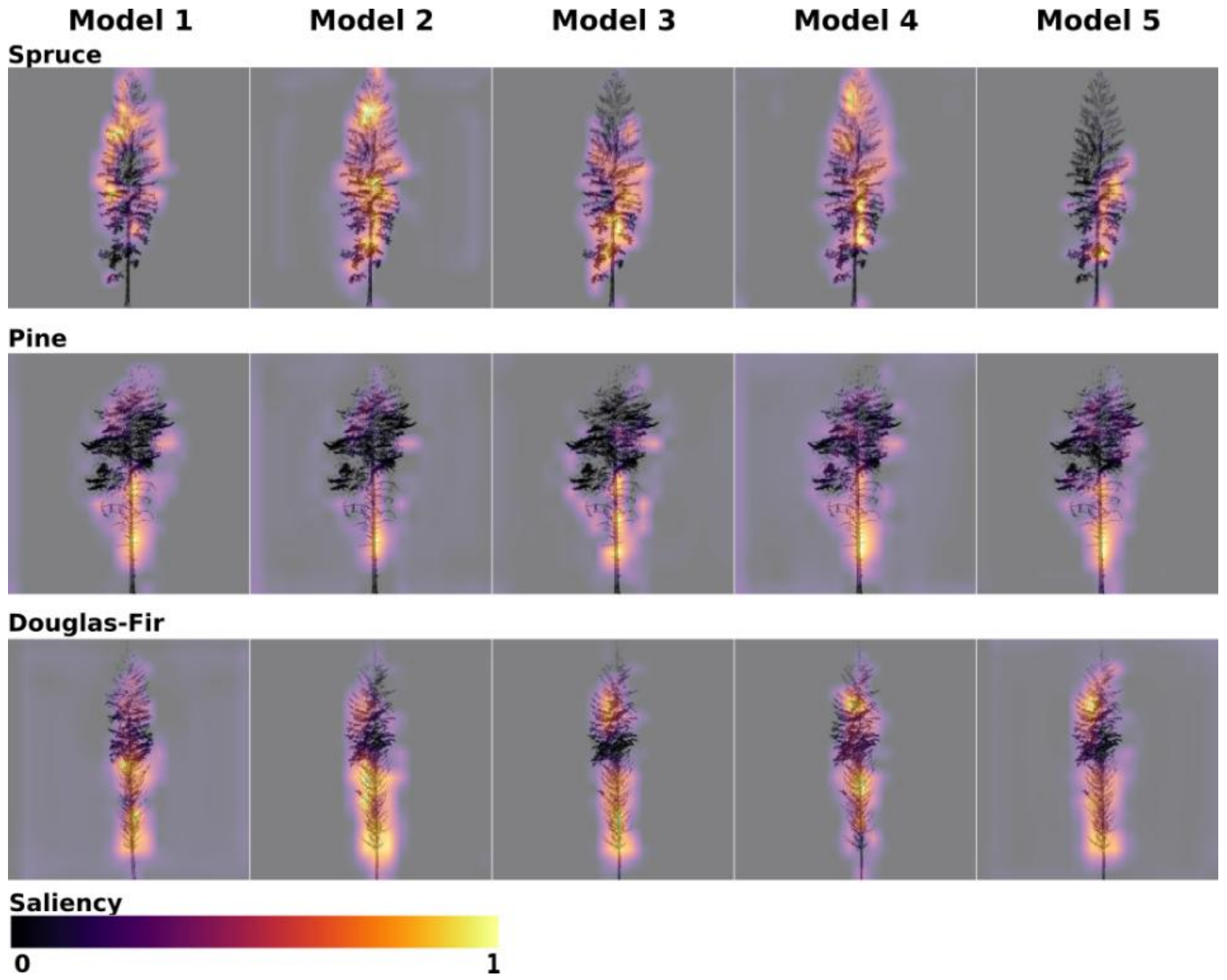

Figure 10 Example saliency maps of the YOLOv8 models trained on the non-perturbed data set of the same side-view projections for Spruce, Pine, and Douglas-fir overlayed with the TLS side-view projections. Layer activations for a species class are depicted in a colour gradient from purple to yellow, where yellow and purple correspond to the regions with high and low saliency, respectively.

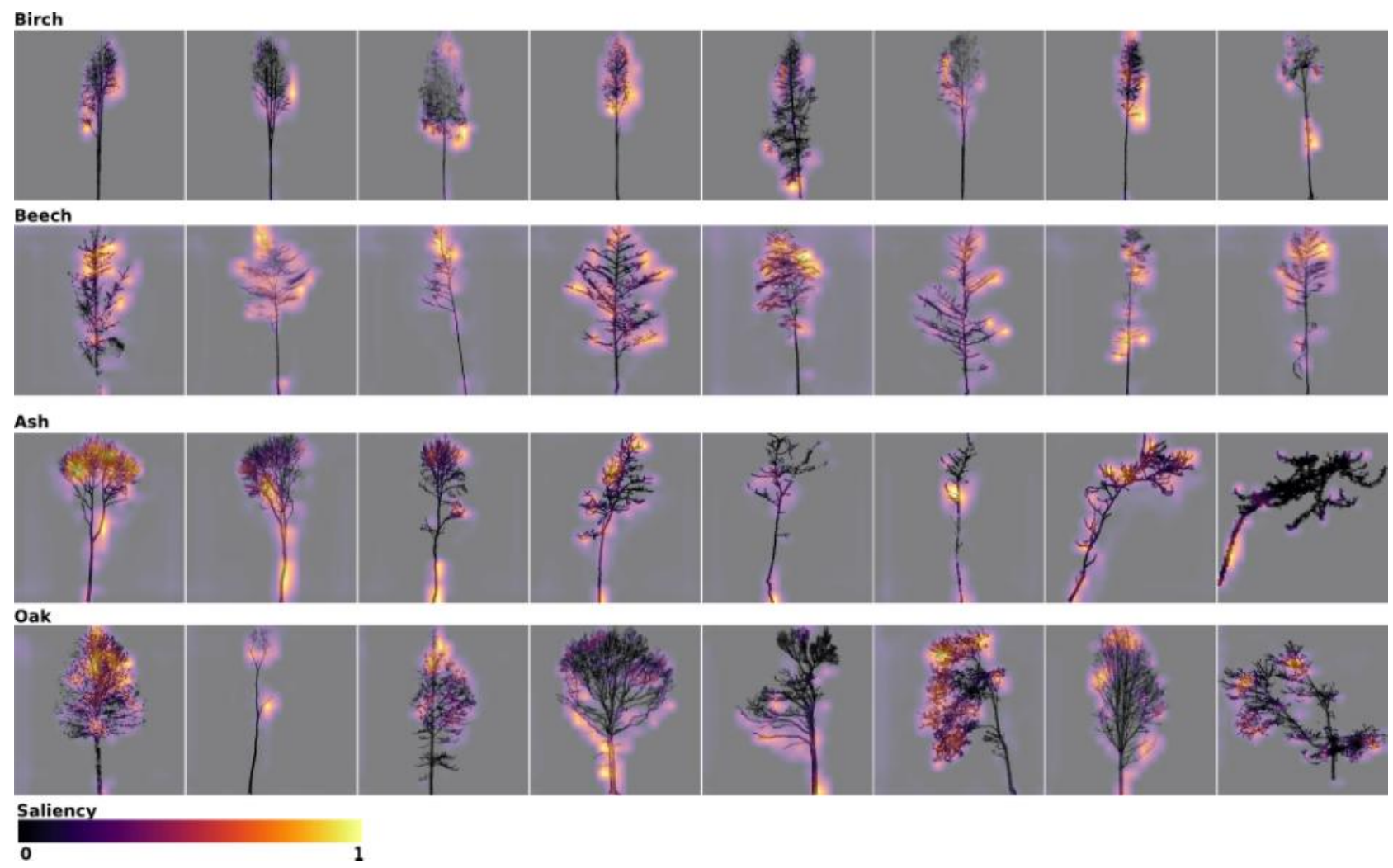

Figure 11 Example saliency maps of model 1 trained on the non-perturbed data set for Birch, Beech, Ash, and Oak overlayed with the corresponding 2D side-view images. Layer activations for a species class are depicted in a colour gradient from purple to yellow, where yellow and purple correspond to the regions with high and low saliency, respectively.

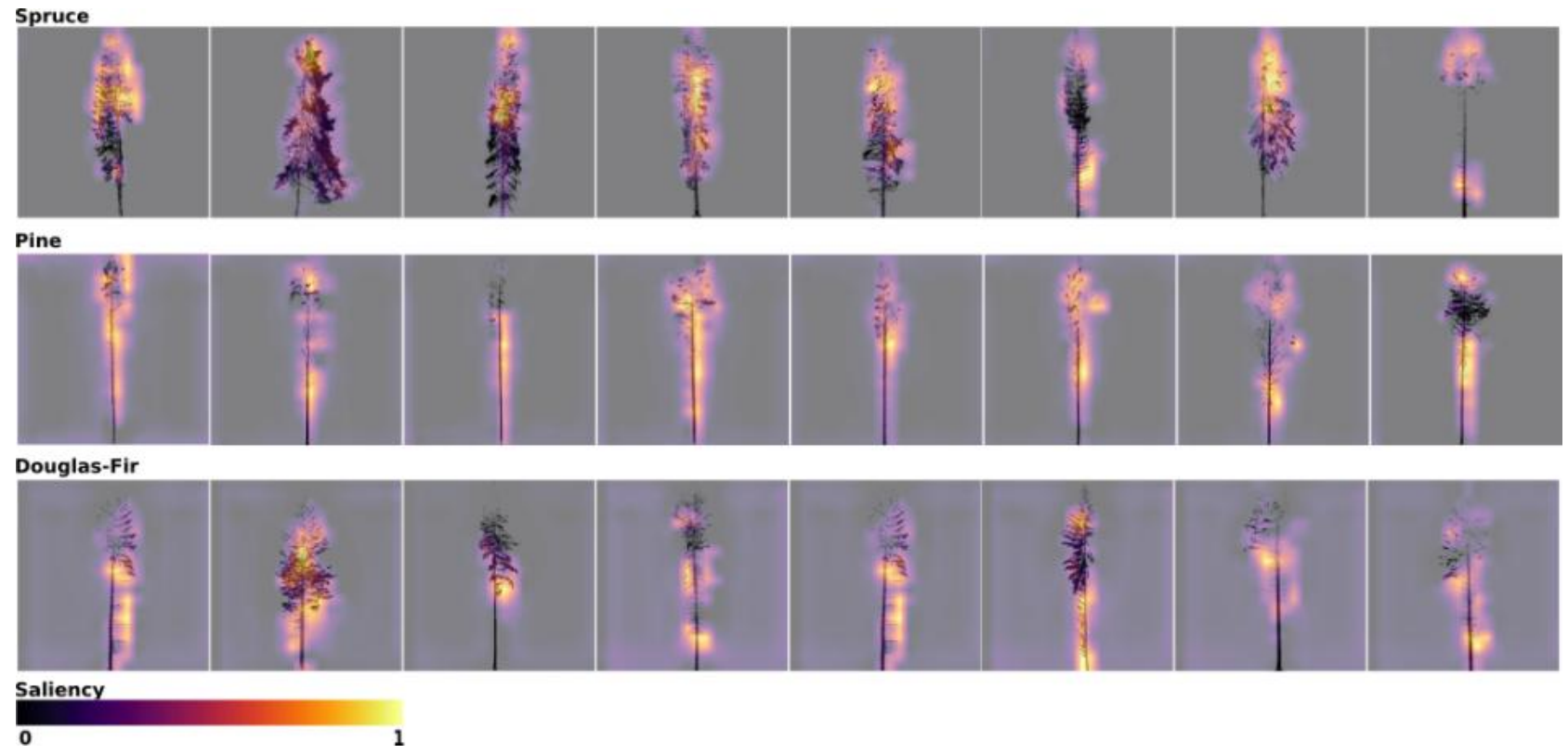

Figure 12 Example saliency maps of model 1 trained on the original non-perturbed data set for Spruce, Pine, and Douglas-fir overlayed with the corresponding 2D side-view images. Layer activations for a species class are depicted in a colour gradient from purple to yellow, where yellow and purple correspond to the regions with high and low saliency, respectively.

The aggregation of the contrastive tree species of all saliency maps of the five models is presented in Figure 13. In general, the models' output logits of broad-leafed tree species are frequently on second and subsequent ranks when other broad-leafed tree species are the target species. When *Spruce* and *Douglas-fir* are the target species the model's output logits for other coniferous tree species rank most frequently on second and subsequent positions. However, the model's output logits for the broad-leafed tree species *Birch* ranks most frequently on the second position when the target species is *Pine*.

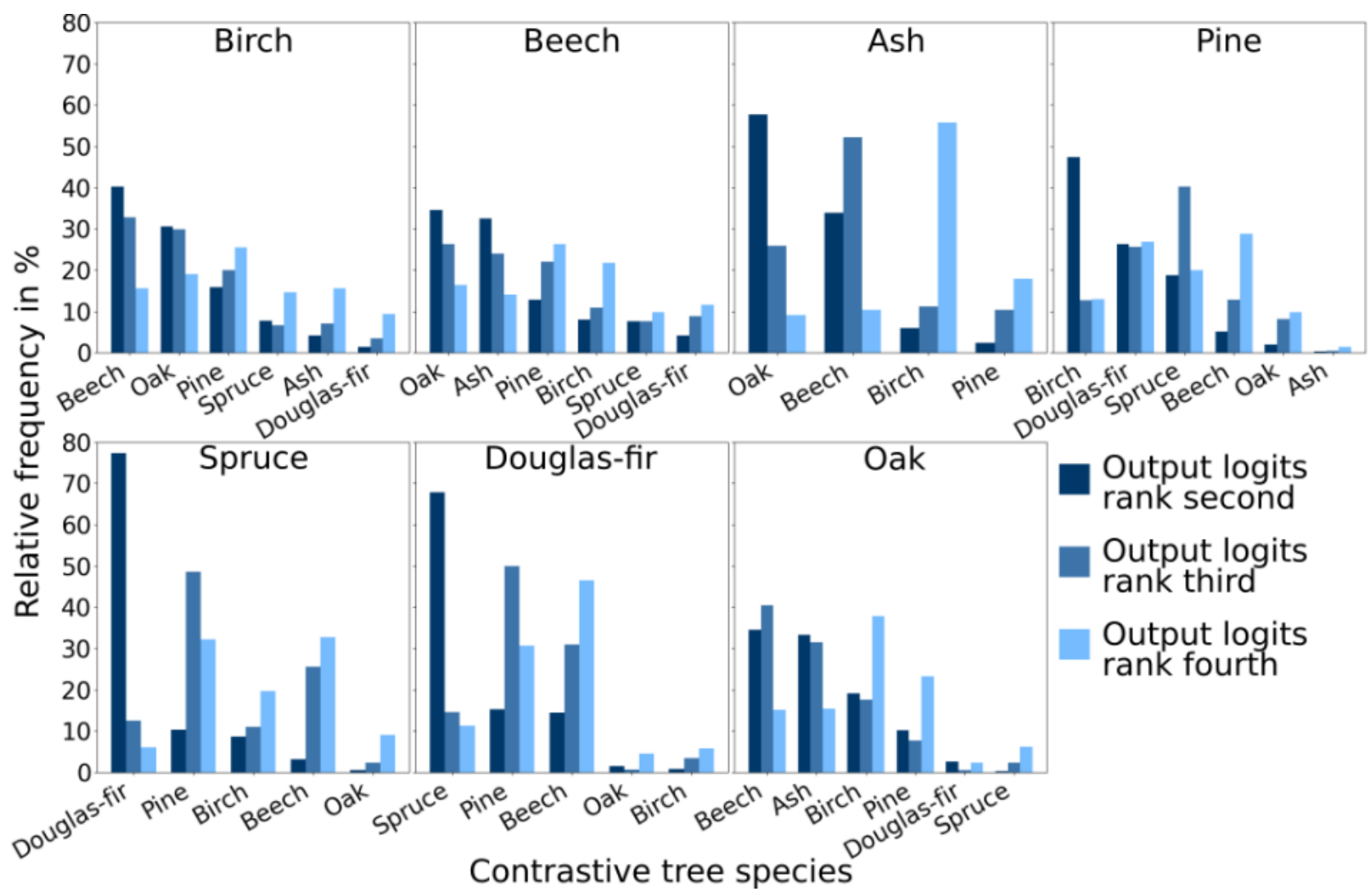

Figure 13 Frequency of species contrastive to a predicted target tree species, of which the output logits rank second, third and fourth behind the output logits of the target tree species. Output logit ranks are aggregated over all Finer-CAM saliency maps and YOLOv8 models trained on the non-perturbed data set.

To evaluate the contribution of regions in the 2D side-view images associated with structural tree features i.e., stem, crown, and crown sub segments, to the classification decisions of the five YOLOv8 models over all tree species, we present the segment-wise arithmetic means and standard deviations of the aggregated ratios of salient pixels of all tree species and models in Figure 14. On average most salient pixels are in the crown segments (68.6 %).

Among the crown sub-segments (crown top, crown middle, crown base) the average ratio of salient pixels is lowest in the crown top segments. On average nearly as many salient pixels of the crown segments are located at the edges of the crown as within crown centres (average ratio of salient pixels in crown edge buffer = 48.1 %).

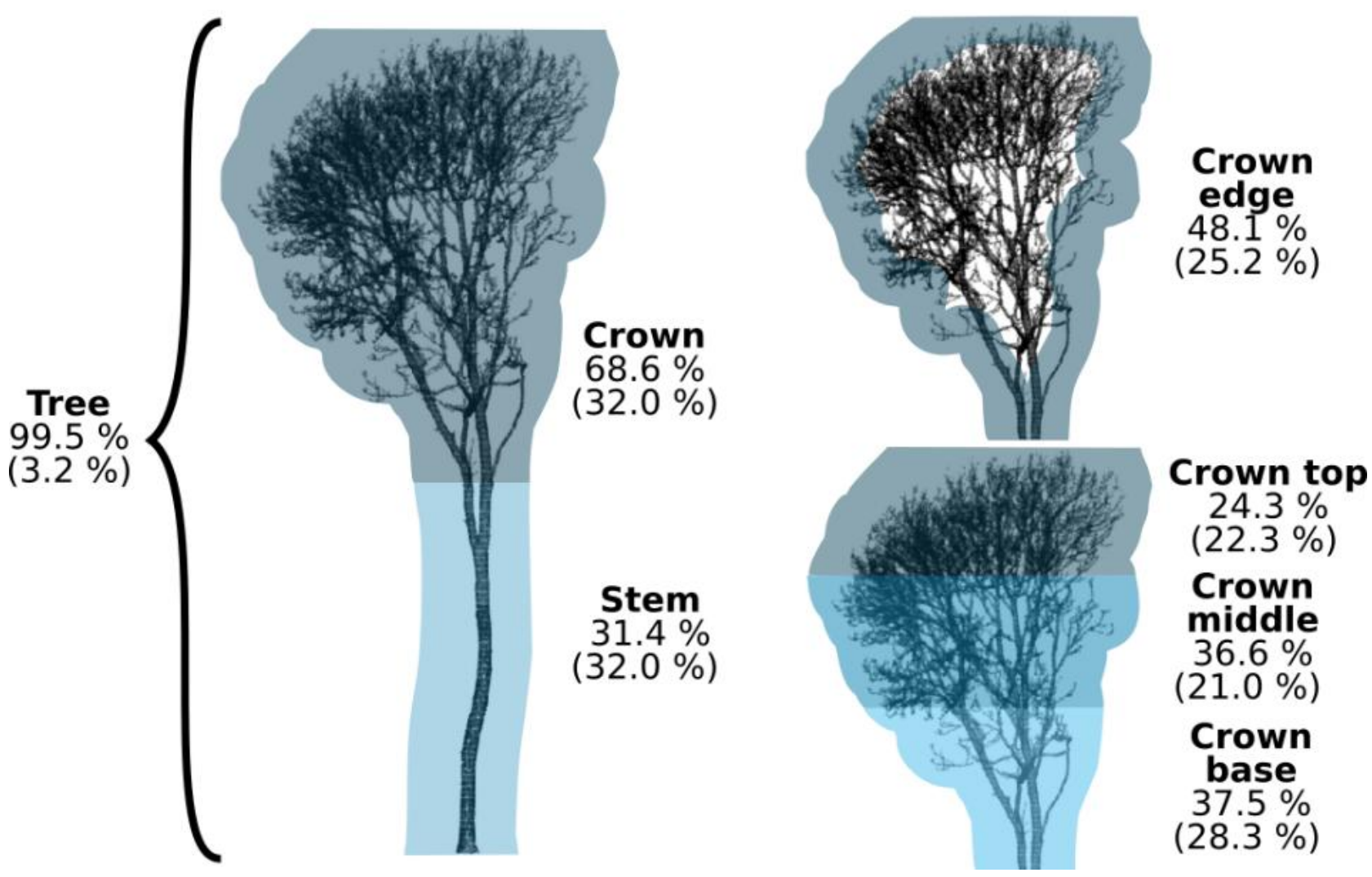

Figure 14 Segment-wise arithmetic means and standard deviations (in brackets) of the ratios of salient pixels associated to image segments associated with structural tree features aggregated over all tree species and YOLOv8 models trained on the non-perturbed data set.

In Figure 15 we present differences and similarities among the tree species concerning the contribution of regions in the 2D side-view images associated with structural tree features to the classification decisions of the models Multiple pairwise comparisons using the Dunn's test with adjusted p-values using the Holm-Bonferroni method show significant differences of the ratios of salient pixels in the stem and crown segments between *Ash*, *Pine*, and *Douglas-fir* and the other tree species. The ratios of salient pixels in the stem segments of *Ash*, *Spruce*, and *Douglas-fir* are significantly higher than of the other tree species.

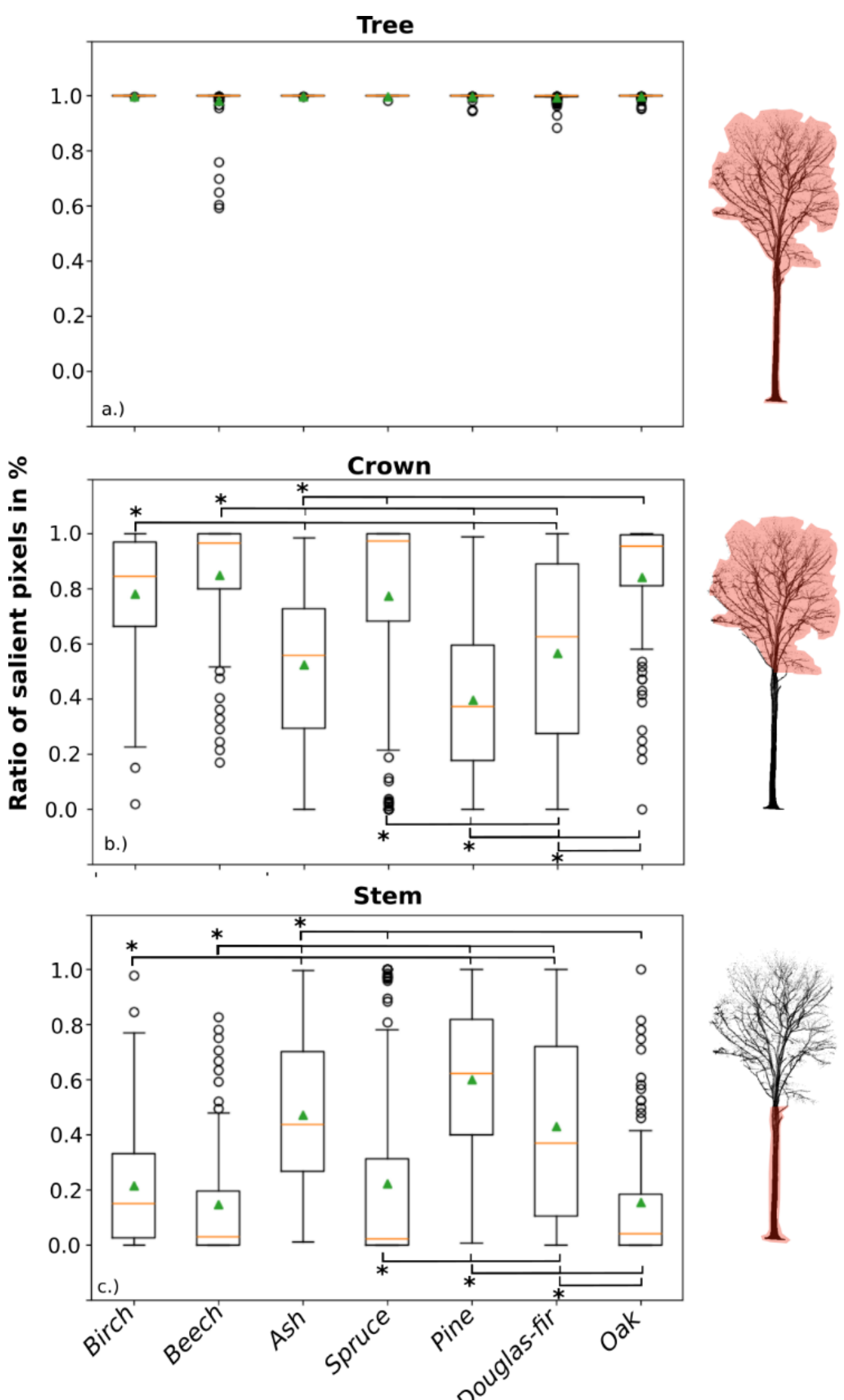

Figure 15 Species-wise box plots of the ratios of salient pixels from 640 saliency maps of the five YOLOv8 models trained on the non-perturbed data set. Saliency maps were generated for a random selection of 128 side-view projections of seven tree species of the test data set. The ratios of salient pixels are associated to tree a.), crown b.) and stem c.) segments. Horizontal lines in boxes represent the medians, green triangles the arithmetic means, the box boundaries the first and third quartile, the whiskers are the first quartile minus 1.5 times the interquartile range and the third quartile plus 1.5 times the interquartile range. The circles show outliers. Brackets below and above the box plots in b.) and c.) indicate significant (alpha = 0,05) differences after p-value adjustment using the Holm-Bonferroni method between species of interest (marked with asterisk) and other species.

The tree species comparisons of the ratios of the salient pixels in the crown sub-segments (crown base, crown middle, crown top, crown edge buffer) averaged across the results of all five YOLOv8 models are presented in Figure 16. Here it shows that the average ratios of salient pixels in the crown middle segments are similar among all tree species except for *Spruce*, which has significantly the highest average ratio of salient pixels in the crown middle segment. Comparing the average ratios of salient pixels of the crown top segments with the ratio of salient pixels of the crown base segments of *Birch*, *Beech*, *Pine*, and *Douglas-fir* it is observable that while the average ratios of the salient pixels of the crown top segments are low, the average ratios of the salient pixels of the crown base segments are high, and vice versa.

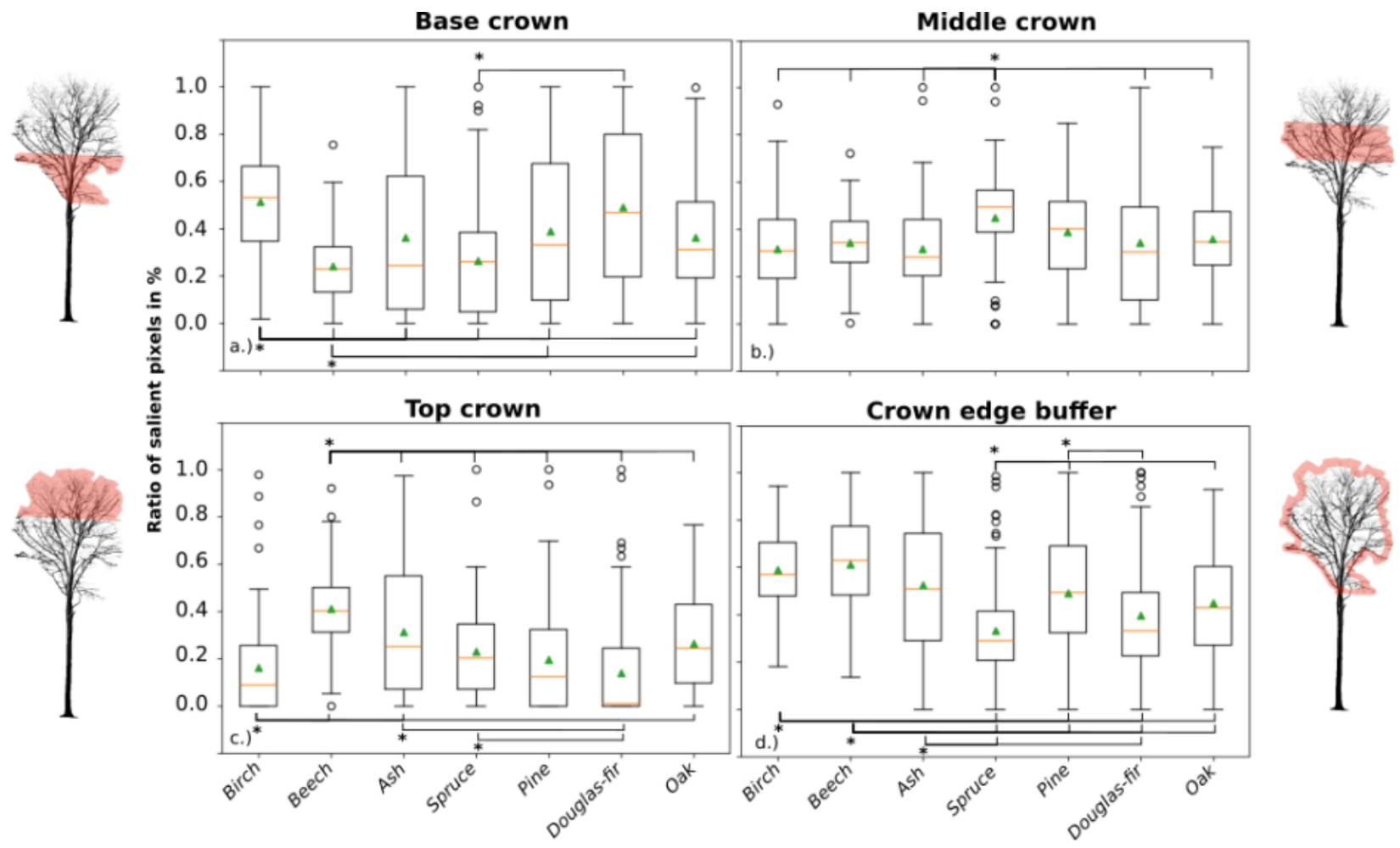

Figure 16 Species-wise box plots of the ratios of salient pixels from 640 saliency maps of the YOLOv8 models trained on the non-perturbed data set. Saliency maps were generated for a random selection of 128 side-view projections of seven tree species of the test data set. The ratios of salient pixels are associated to crown sub-segments crown base a.), crown middle b.), crown top c.), and crown edge buffer d.), segments. Horizontal lines in boxes represent the medians, green triangles the arithmetic means, the box boundaries the first and third quartile, the whiskers are the first quartile minus 1.5 times the interquartile range and the third quartile plus 1.5 times the interquartile range. The circles show outliers. Brackets below and above the box plots in b.) and c.) indicate significant (alpha = 0,05) differences after p-value adjustment using the Holm-Bonferroni method between species of interest (marked with asterisk) and other species.

### *Experiments image perturbation*

We trained five individual models on 2D side-view images representing binary tree shapes (*experiment shape*). The highest overall accuracies during training were achieved after 27, 27, 34, 40, and 49 epochs for the five models. On average the models achieved a macro F1-score of 78.8 % (SD = 1.3 %), a macro recall of 83.2 % (SD = 1.1 %), and a macro precision of 76.5 % (SD = 1.1 %).

In the *tree structure visibility reduction experiment* we trained 35 models on perturbed 2D side-view images which resemble binary tree representations with or without reduced image quality. To achieve quality reduction, we downsampled the 2D side-view images to sizes of 320, 160, 80, 40, 20, 10 pixels which correspond on average over all species and 2D side-view images of the training/validation and test data sets to effective spatial resolutions of 0.06 m, 0.11 m, 0.23 m, 0.46 m, 0.93 m, 1.87 m, respectively. The average spatial resolution of the non-perturbed 2D side-view images of the training/validation and test data sets is 0.03 m. The macro F1-scores of each model and experiment averages are depicted in Figure 17.

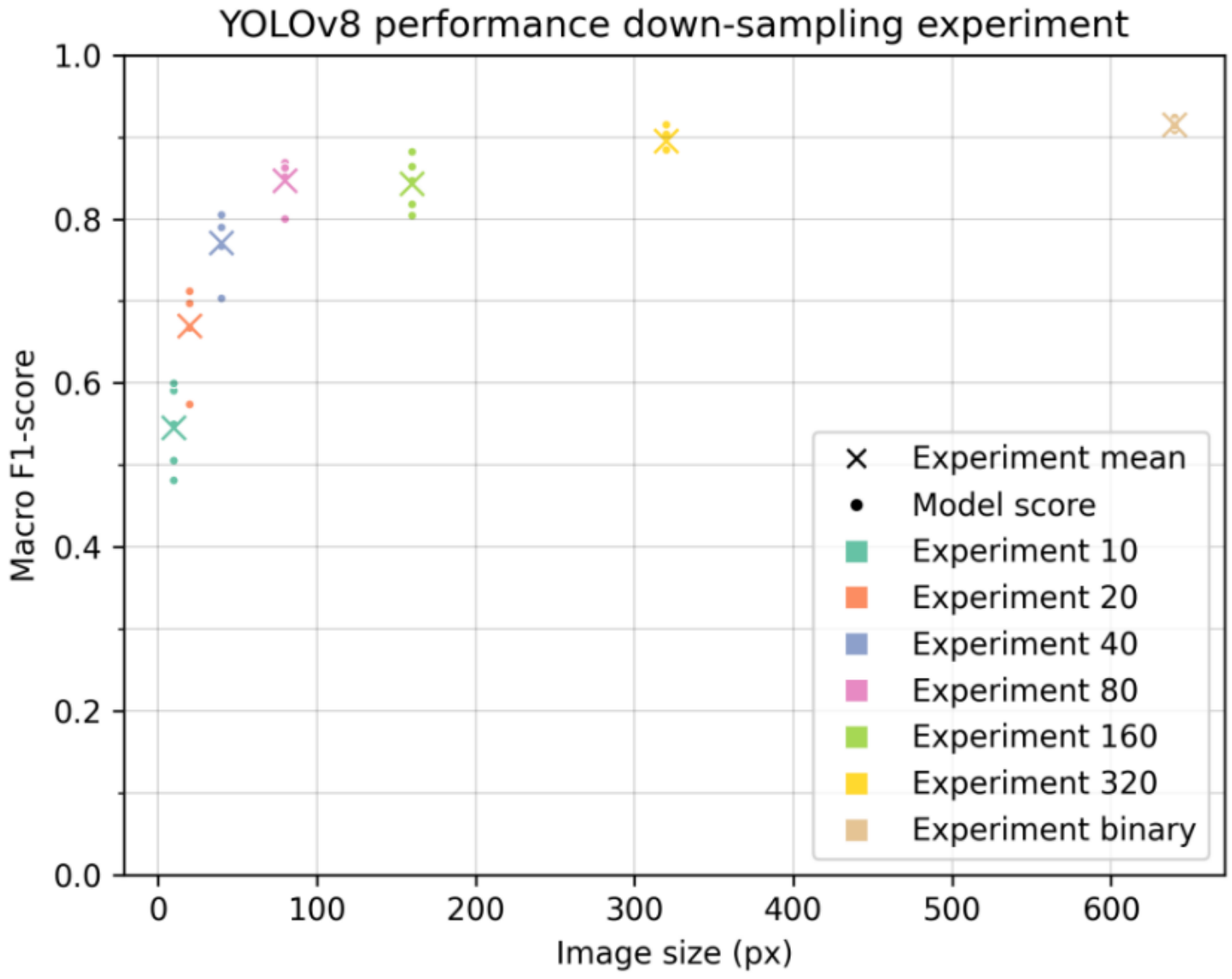

Figure 17 Performances (F1-scores) of YOLOv8 models of the image perturbation experiments. We trained 35 YOLOv8 models on perturbed 2D side-view images in which we transformed the original non-perturbed 2D side-view projections into binary images and downscaled them to image sizes of 10 px, 20 px, 40 px, 80 px, 160 px, and 320 px to reduce visibility of internal tree structures.

## Discussion

The five YOLOv8 models trained and tested on the non-perturbed data set show high classification accuracies (96 %) across all tree species. However, although we aimed to increase generalisability by including only tree species for which TLS data is provided by at least three different data sources in the FOR-species20k data set, the classification accuracies might be optimistic due to spatial autocorrelation within the data sources, as TLS 3D point clouds of individual trees from a data source are collected in spatial proximity to each other. The models most frequently confuse coniferous with other coniferous and broad-leafed with other broad-leafed species, which is relatable from our perspective and generally in accordance with the assessment of each target species' most contrastive tree species (see Figure 13).

Our experiments indicate that Finer-CAM saliency maps can be considered faithful in identifying image regions with a high explanatory power to the classification decisions. The five models trained independently on the non-perturbed data set remain confident in correctly identifying the target tree species even when only 20 % of most relevant pixels —determined in terms of Finer-CAM saliency attribution to species discrimination— are presented to the models. When more than 80% of pixels with lowest saliency are removed, the average model confidence in assigning the target species class decreases rapidly across all species and models. These model behaviors are not observed when applied to 2D side-view images in which pixels were removed based on random saliency maps, which underscores the explanatory power of Finer-CAM. The results of the faithfulness experiment suggest that only a fraction of data may be needed for YOLOv8 to discriminate tree species correctly. This could have implications to future conceptions of CNN-based tree species classification approaches in a sense that it may be feasible to conceptualize approaches in which only detailed TLS 3D point clouds —or 2D projections of them— representing tree features with the highest global contribution to a model's classification decisions rather than representations of entire trees are used. As a consequence, this may increase the efficiency in assembling training data sets, since multiple representations of tree features with high contribution to classification decisions could be sampled from an individual tree TLS 3D point cloud, which could reduce the number of individual tree scans necessary for the creation of a training data sets. DetailView which is benchmarked in Puliti et al. (2025) is a latest approach using detailed representations of stem features from TLS 3D point clouds.

Our results suggest that 2D side-view images of TLS 3D point clouds describe the habitus of trees, in the sense of the internal tree structure and branching patterns, which can be used by YOLOv8 models to discriminate tree species. In the image perturbation experiments YOLOv8 shows the highest performance when applied to 2D side-view images in which internal tree structure is visible (see average F1-score of YOLOv8 models when applied to binary 2D side-view images Figure 17), indicating the capabilities of YOLOv8 to leverage detailed point cloud representations, which is also emphasized by Puliti et al. (2025). Performances (F1-scores) of YOLOv8 models remain high until the average effective spatial resolution of the 2D side view images is below approximately 0.5 meters (downsampling to image size of 40 px), which resembles a point from which visibility of internal tree structure like individual branches is limited. Contrary, our experiments show YOLOv8s capabilities of identifying tree species correctly even when applied to 2D side-view images with reduced level of detail and thus visibility of internal tree structure (see Figure 17). YOLOv8 models even achieve sound performances when trained and tested on 2D side-view images representing the tree shape —which is the filled-out area within the contours of a tree— (F1-score = 78,8 %), indicating YOLOv8s capabilities to leverage tree shape for tree species classification. However, this still constitutes a decrease of model performance of about 15 % in comparison to the YOLOv8 models applied to the non-perturbed 2D side-view images in which internal tree structure is visible. These results show the capabilities of YOLOv8 to also leverage information from sparse data and tree shape which indicate potentials of the adoption of a YOLOv8-based tree species classification approach in the context of using sparse 3D point clouds from other data collection approaches like airborne laser scanning for tree species classification. However, the results indicate the importance of visibility of internal tree structure in 2D side-view images when the goal is to achieve high model performance. Additionally, our results suggest that the contribution of point cloud density that translates into pixel values of the 2D side-view images to the classification decisions of the models can be regarded marginal, since the models applied to binary 2D side-view images perform similar as the models applied to the non-perturbed gray-scale 2D side-view

The analysis of saliency maps for the five models trained and tested on the non-perturbed 2D side-view images further support the conclusions drawn from the perturbation experiments and reveal additional insights into the classification decisions of the models. On average, segments in the 2D side-view images corresponding to the tree itself contribute substantially to tree species discrimination, with an average of 99.5 % of salient pixels located within the tree segment, across all five YOLOv8 models. This result confirms an expected behavior of the models: Regions in the 2D side-view images corresponding to tree features contribute to the classification decisions of the models more than i.e., image backgrounds. Our results show that regions associated with tree crowns contribute most to the classification decisions for differentiating tree species (see Figure 14). The pronounced focus on crown regions can be particularly observed for *Birch*, *Beech*, *Spruce*, and *Oak*. A visual interpretation of the saliency map examples of these tree species suggests, that features identified by Finer-CAM as having high contributions to the model classification decisions correspond to branches in the tree crowns.

This may indicate that YOLOv8 has the capability to learn species-specific branching patterns, particular of finer branches, to differentiate tree species, which would represent a plausible mechanism from a biological perspective. We find that, for *Birch*, *Beech*, and *Oak* around 50 % of the salient pixels within the crown segments are in the crown edge buffer segments. As finer branches are particularly discernable at the edges of tree crowns, this result further supports the interpretation that local tree structures visible in 2D side-view images derived from TLS 3D point clouds contribute to the tree species classification of YOLOv8. Our results further indicate that for *Birch* image features in the crown base segments and for *Beech* image features in the crown top segments contribute most to the classification decisions. Visual interpretation of the saliency maps for both tree species reveals that most salient regions are located at regions associated with branches. For *Birch* trees branches are more clearly visible at the base and towards the edges of the crown than at the top of the tree crowns. For *Beech* trees branch structures are often visible along the entire length of the crown, including the top and particularly at the edges of the tree crowns. These observations can be interpreted in a way that they indicate the models' may be capable of learning species specific branch features, which would support the above presented interpretation. Contrary, these results may suggest that the model discriminate both tree species based on the location (height along the vertical axis) at which features associated with branches are visible in the 2D side-view images.

Across all models, the average proportion of salient pixel in crown segments is higher than in stem segments for nearly all tree species (except *Pine*). Nevertheless, our results indicate that features in stem regions also contribute to the models' classification decisions. For *Ash*, *Pine*, and *Douglas-fir*, Finer-CAM salient regions are on average more frequently located at stem-associated features. Visual interpretation of saliency maps for *Pine* and *Douglas-fir* shows that features associated with branches attached to the stem show frequently a high saliency (see Figure 10 and Figure 12). Both tree species tend to retain dead branches at the stems since their low self-pruning rates (Schönauer et al. 2021, Kothari et al. 2025). This observation may indicate the potential of the models to learn stem features specific to *Douglas-fir* and *Pine*. For *Ash* trees, all five models focus on bend parts of the stem in 64.4 % of the analyzed saliency maps. This explains the high contribution of stem segments to the classification decisions (see Figure 9 and Figure 11). Bends in stems can occur in any tree species as a result of tree growth in response to external factors such as changes in light conditions, wind, and snow. Consequently, the occurrence of stem bends cannot be considered specific only to *Ash trees*. This result may indicate that the models are learning a data set specific feature namely that stem bends are common for *Ash* in the data set but not for its most similar tree species, rather than structural features specific to *Ash*. This effect may be reinforced by the low number of Ash trees in the data set and could lead to a high failure rate when the models are applied to unseen data. The evaluation of the contrastive species (see Figure 13) indicates that the model's output logits for certain tree species are similar, which we would also judge to be alike based on their architectural traits. Among the five models trained and tested on the non-perturbed data set the output logits of *Spruce* and *Douglas-fir* are most frequently on the second rank when one or the other scores highest. The output logits of *Pine* frequently ranked third when either *Spruce* or *Douglas-fir* was the target class. From our perspective —while all are coniferous— only *Spruce* and *Douglas-fir* show similarities in tree structure. The output logits of Broad-leafed species are most frequently on the fourth rank for *Spruce* and *Douglas-fir*, which is expectable since in we only consider three coniferous species. For *Pine* the output logits of *Birch* rank most frequently on second position and the output logits of Beed on fourth position.

Furthermore, the output logits of *Pine* rank most frequently on fourth position when the target species is *Birch* or *Beech.* From our perspective this is relatable since the habitus of *Pine* can show similarities to broad-leafed tree species. The output logits of broad-leafed tree species rank most frequently on the second position when broad-leafed species are the target species, which is reasonable from our perspective.

It is difficult to directly compare our results to other studies investigating saliency maps in the context of tree species classification using individual tree TLS point clouds, since these studies focus on mostly different tree species, only present saliency maps for some example instances, and analyse them only visually (e.g., Xi et al. 2020, Liu et al. 2022), which is prone to be affected by human biases (Mohseni et al. 2021, Bertrand et al. 2022). Additionally, these studies applied tree species classification models performing directly on the individual tree TLS point clouds and present saliency maps projected onto point clouds or so-called critical points that contribute most to the learned features in the used CNNs. Liu et al. (2022) report that critical points primarily appear in the tree crowns. This observation is consistent with our findings, which indicate that regions in the 2D side-view images associated with the tree crowns contribute most to the classification decisions of the models. Xi et al. (2020) conclude that crown features contributed to the classification of four (*Cottonwood*, *Maple*, *Spruce*, *White Spruce*) of their tested nine tree species. For the other tested tree species (*Lodgepole Pine*, *Red Pine*, *Scots Pine*, *Birch*, and *Aspen*), the stem appeared to be more important for a correct classification, which is also in line with our results showing a high contribution of regions in the 2D side-view images associated with stem and crown features for Pine and *Spruce*, respectively. In regard to *Birch* their results differ from ours, which is difficult to interpret, because of the differences in methodology between our study and the study conducted by Xi et al. (2020).

Our proposed approach, which links Finer-CAM saliency map regions with high contributions to model classification decisions to structural tree features (stem, crown, crown sub-compartments), facilitates better interpretability of CNN-based tree species classification models using 2D side-view projections of TLS 3D point clouds. However, the results highlight limitations in the conclusions that can be drawn from our approach. Our results may be negatively affected by errors introduced during the application of image partitioning (generation of segments) and the allocation of salient pixels to image segments. Firstly, the visual determination of the crown base used to differentiate crown and stem segments can be regarded as subjective, which could have influenced the allocation of salient pixels to these segments. However, this source of uncertainty is considered minor given that the visual determinations were independently verified. Secondly, discrepancies introduced during the tree contour generation process may also have impacted the analysis results in a way that salient pixels may be wrongly associated to segments. In 2D side-view images in which tops of the tree crowns are under-represented in the 2D side-view images due to TLS inherent point cloud characteristics (reduced point density in tree tops), the automated process for tree contour detection has resulted in the creation of crown segments, in which the tops are cut off. This issue occurred more frequently for coniferous trees than for broadleaved trees, since in the used data set top crowns are better represented in 2D side-view images of broadleaved trees. However, in total not many generated crown segments were affected. An additional limitation is imposed by the application of Finer-CAM saliency maps, since their use is constrained, as these maps provide only visual explanations rather than a direct assessment of tree feature importance, introducing the potential for interpretative bias.

Although we sought to reduce this bias by systematically linking visually salient regions in 2D side-view images —identified by Finer-CAM as contributing to the classification decisions of the YOLOv8 models— to structural tree features (stem, crown, crown sub-compartments), our approach does not allow for a final assessment of which specific tree features drive the models' classification decision. However, since our results indicate faithfulness of Finer-CAM saliency maps, we consider our approach feasible in identifying general trends of the contributions of structural tree features to the models' classification decisions. Additionally, the sparse availability of methods from the field of XAI suitable for assessing the direct contribution of features to the classification decisions of tree species classification models renders our proposed suitable to better understand how a CNN-based tree species classification approach assigns 2D side-view projections of TLS 3D point clouds of individual trees to a target species class, until more advanced methods for assessing tree feature importance in CNN-based tree species classification models are available. Consequently, our results should not be interpreted as an absolute statement on the global contribution of tree features for tree species classification but rather as an indication of which features are most accessible to the models under the chosen study design. The visibility of tree features in 2D side-view images is linked to the design of these. Thus, the models can only learn tree features that are present in the 2D side-view images. This may be further pronounced because of the applied scaling for projecting the TLS point clouds onto 2D images, which is linked to tree size (height). Because of the scaling tree size is directly translated into varying spatial resolutions of the 2D side-view images. Thus, features of smaller trees are likely to be represented in finer detail than for larger trees, possibly affecting which tree features are learned by the models and overall model performance. Consequently, spatial resolution could contribute to the classification decisions of the models in a way that they learn to differentiate tree species by means of tree size, or that they learn different features for classifying smaller and larger trees. While tree heights of the tree species in the used data set are to some degree unequally distributed (see Figure 1, for example *Douglas-fir* against the others), we regard the impact of these differences on the classification decisions of the model to be marginal in a sense that we do not expect them to be a major driver for the discrimination of tree species, since there is overlap between the tree species height distributions. However, a certain contribution of tree size to the classification decisions of the models cannot be dismissed, since our proposed approach does not allow for its assessment. Thus, the effects of spatial resolution on which features are learned by the models and its contribution to classification decisions of the models should be investigated further in subsequent studies. Accounting for these potential effects of spatial resolution was infeasible in this study because of the relatively low number of individual tree TLS scans for some tree species (*Ash, Oak, Birch, Douglas-fir*) in the data set. This can be regarded as a data set limitation, which further highlights the need for comprehensive public TLS data sets consisting of trees in various developmental stages from different regions and for which TLS point clouds are collected with various scanners from different working groups. While the relatively low number of individual tree TLS scans of some tree species included in the data set may weaken the robustness of the conclusions drawn from the species-wise analysis of the saliency maps, the fact that we analysed saliency maps of five independently trained models allows to distil a general trend about structural tree feature contribution to classification decisions and thus helps to enhance model interpretability. In summary, a final assessment of the magnitude of tree feature importance to classification decisions of the models remains difficult, but our presented approach represents a new path to address the need for explanations of model classification decisions in the field of tree species classification using TLS scans. Insights from this study highlight the need for future research that i.) establishes approaches which allow for a direct assessment of tree feature importance similar to other machine learning techniques like in random forest (Brown et al. 2025) ii.) quantitatively assesses tree feature contributions to classification decisions of other CNN-based tree species classification approaches, iii.) investigates the occurrence of data set artifacts of publicly available TLS data sets and their effects on the classification decision of CNN-based tree species classification models, and iv.) analyses the generalizability of CNN-based tree species classification models trained on public benchmark data sets by application on fully independent test data sets. We argue that findings in these fields have the potential to create sophisticated quality standards, which include the definition of field and TLS data processing protocols, for TLS data sets used for training and benchmarking of tree classification models enhancing the performance stability of developed tree species classification approaches and user's trust in them.

## Conclusion

In this study we present a novel framework in which we link regions of Finer-CAM saliency maps with a high contribution to the classification decisions of YOLOv8 models to regions in 2D side-view images associated with structural tree features (e.g., stem, crown, crown sub-segments) to gain more insights into the classification decisions of the models. Our results indicate that Finer-CAM saliency maps can be considered faithful in identifying most discriminative image regions that distinguish the target species classes from similar species classes, rendering our proposed approach suitable to increase model interpretability. For the seven tree species considered (*Birch*, *Ash*, *Beech*, *Spruce*, *Pine*, *Douglas-fir*, *Oak*) our results indicate that image regions associated with tree crowns contribute most to the classification decisions of the models. This observation is particularly pronounced for *Birch*, *Beech*, *Spruce*, and *Oak*. For *Ash*, *Pine*, and *Douglas-fir*, image features in stem segments contribute most for the decisions of the models. Visual interpretation of saliency maps for *Spruce* and *Douglas-fir* indicates that regions in the 2D side-view images corresponding to branches attached to stems contribute to the classification decisions of the models. Consequently, our results suggest an ability of the models to leverage structural features specific to the species, such as the presence of branches retained or characteristics of the internal crown structure, in distinguishing between different tree species. Additionally, our results indicate that visibility of tree structures in the 2D side-view images affects YOLOv8 model performance positively, suggesting the capabilities of YOLOv8 to leverage detailed point cloud representations. For *Ash* the models focus on average in 64.4 % of the 2D side-view images on bends in tree stems. Since bent tree stems are not a trait exclusive to *Ash* our results may indicate a data set specific artifact, where bent stems are predominantly associated with *Ash* trees in the used data set. This may limit generalizability of the models when applied to unseen data. Our results should be interpreted as an indication of which image features associated with tree structures are most accessible to the models under the chosen study design. In this regard the presented approach can help to enhance interpretability of a model's classification decisions by indicating its limitations and potential biases (shortcut learning). Our results further highlight that examining how deep learning models perform the task of tree species classification using TLS data in more depth than only reporting performance results can be a key-step to enhance the quality of data sets used for development and benchmarking of classification models, which can benefit the performance and stability of classification models.

## Funding

The research leading to these results has received funding from the European Union Horizon Europe Research & Innovation programme under the Grant Agreement no. 101056907 (PathFinder). Further it was part of the Center for Research-based Innovation SmartForest: Bringing Industry 4.0 to the Norwegian forest sector (NFR SFI project no. 309671, smartforest.no).

## Conflict of interest statement

None declared.

## Data availability statement

The data set underlying this article are available in the FOR-species20K data set repository, at https://doi.org/10.5281/zenodo.13255197. The code of the explanation framework and the tree species classification approach can be downloaded from https://doi.org/10.5281/zenodo.18958943.

## Author contributions

Adrian Straker developed the ideas, designed the methodology, performed the data analysis, and led the manuscript writing. Marco Zullich provided substantial knowledge in the field of XAI and contributed the section *Finer-CAM description*. Nils Nölke and Paul Magdon supported the development of the research design. Johannes Breidenbach supported in the interpretation of the results. Max Freudenberg developed the 2D side-view projection approach of 3D TLS point clouds. Christoph Kleinn supported the statistical analysis. Stefano Puliti supported the dataset compilation. All authors contributed critically to the drafts.